\documentclass{article}
\usepackage{amssymb}
\usepackage{graphicx}
\usepackage{tikz}
\usetikzlibrary{arrows.meta, positioning}
\usepackage{amsmath}
\usepackage{algorithm}
\usepackage{algpseudocode}
\usepackage{listings}
\usepackage{pifont}

\newcommand{\cmark}{\ding{51}}
\newcommand{\xmark}{\ding{55}}

\lstdefinestyle{python}{
    language=Python,
    basicstyle=\ttfamily\footnotesize,
    keywordstyle=\color{blue}\bfseries,
    stringstyle=\color{red},
    commentstyle=\color{green!50!black},
    showstringspaces=false,
    frame=single,
    breaklines=true,
    tabsize=4,
}

\usepackage{colortbl}
\usepackage{xcolor}
\PassOptionsToPackage{table,xcdraw}{xcolor}
\usepackage{booktabs}
\usepackage{multirow}
\usepackage{bm}
\usepackage{subcaption}
\usepackage{tabularx}

\usepackage[final]{corl_2025} 

\title{First Order Model-Based RL \\ through Decoupled Backpropagation}

\author{
  Joseph Amigo$^{*12}$, Rooholla Khorrambakht$^{1}$,
  \\ \textbf{Elliot Chane-Sane$^{2}$, Nicolas Mansard$^{23}$, Ludovic Righetti$^{13}$}
  \\ $^{1}$Machines in Motion Laboratory, New York University, USA
  \\ $^{2}$LAAS-CNRS, Universit\'e de Toulouse, CNRS, Toulouse, France
  \\ $^{3}$Artificial and Natural Intelligence Toulouse Institute, Toulouse, France
  \\ {\texttt~\url{https://machines-in-motion.github.io/DMO/}}
}

\begin{document}
\renewcommand{\thefootnote}{\fnsymbol{footnote}}
\footnotetext[1]{Correspondence to \texttt{joseph.amigo@nyu.edu}}
\renewcommand{\thefootnote}{\arabic{footnote}}
\maketitle


\vspace{-1.0cm}
\begin{abstract}
    There is growing interest in reinforcement learning (RL) methods that leverage the simulator's derivatives to improve learning efficiency. While early gradient-based approaches have demonstrated superior performance compared to derivative-free methods, accessing simulator gradients is often impractical due to their implementation cost or unavailability. Model-based RL (MBRL) can approximate these gradients via learned dynamics models, but the solver efficiency suffers from compounding prediction errors during training rollouts, which can degrade policy performance. We propose an approach that decouples trajectory generation from gradient computation: trajectories are unrolled using a simulator, while gradients are computed via backpropagation through a learned differentiable model of the simulator. This hybrid design enables efficient and consistent first-order policy optimization, even when simulator gradients are unavailable, as well as learning a critic from simulation rollouts, which is more accurate. Our method achieves the sample efficiency and speed of specialized optimizers such as SHAC, while maintaining the generality of standard approaches like PPO and avoiding ill behaviors observed in other first-order MBRL methods. We empirically validate our algorithm on benchmark control tasks and demonstrate its effectiveness on a real Go2 quadruped robot, across both quadrupedal and bipedal locomotion tasks.
\end{abstract}



\begin{figure}[h!]
    \centering
    \includegraphics[width=1.0\linewidth]{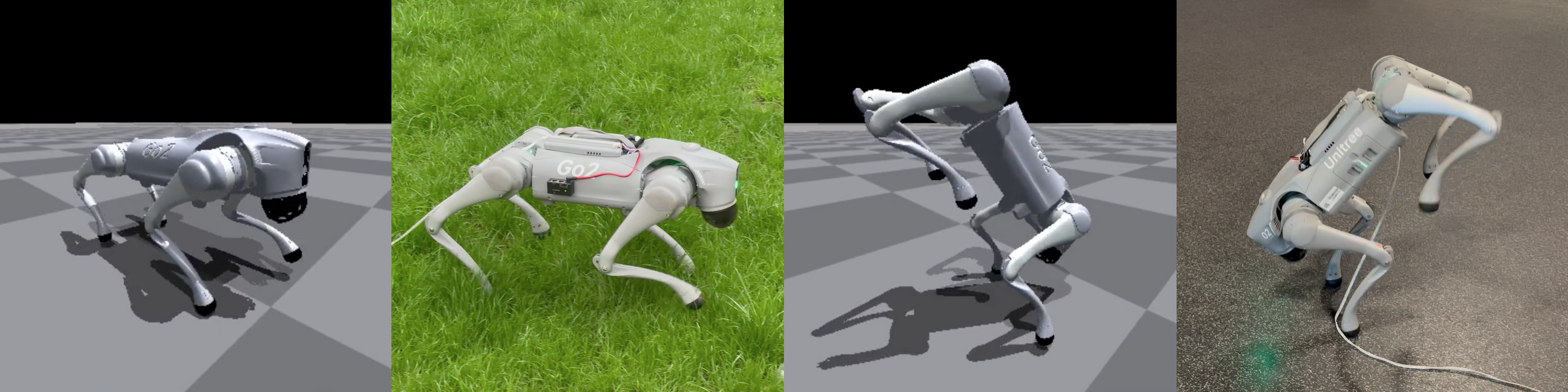}
    \caption{Go2 Walking on four and two legs using policies optimized with DMO. 
    } 
    \label{fig:banner}
\end{figure}
\vspace{-0.5cm}
\section{Introduction}

Reinforcement learning (RL) has led to impressive results in robotics, from agile quadrupedal \cite{cheng2023parkour, hoeller2024anymal, zhuang2023robot, chane2024cat, chane2024soloparkour} and humanoid \cite{radosavovic2024real, zhuang2024humanoid} locomotion to dexterous manipulation \cite{handa2023dextreme, allshire2022transferring}. Popular model-free RL algorithms such as DDPG \cite{lillicrap2019continuouscontroldeepreinforcement}, SAC \cite{haarnoja2018soft, haarnoja2019softactorcriticalgorithmsapplications}, or PPO \cite{schulman2017proximalpolicyoptimizationalgorithms} have powered many of these results, offering generality and strong asymptotic performance.
However, deep RL remains sample-inefficient. Most successful applications rely on massive simulation throughput, leveraging GPU-accelerated environments like Isaac Gym \cite{makoviychuk2021isaac, rudin2022learning} or MuJoCo \cite{todorov2012mujoco} to run thousands of rollouts in parallel. This trend has become central to scaling RL in high-dimensional robotics settings.

Differentiable simulators \cite{freeman2021brax, xu2021accelerated, xing2024stabilizing} have attracted a growing interest, as they allow computing policy gradients directly through physics. These methods often improve learning speed and numerical stability by enabling more informative updates. In a similar spirit, we propose to exploit the additional information of derivatives to enhance learning efficiency. Yet, building fully differentiable simulators remains challenging as contact dynamics are frequently only piecewise differentiable, and implementing gradients for complex environments remains time-consuming and error-prone.

Model-based RL (MBRL) provides an appealing alternative. Rather than relying on simulators that use explicit physical models, MBRL learns a model of the environment directly from data. The result is a differentiable dynamics model that enables both fast GPU-based rollouts and first-order optimization. 
Several recent works have shown that MBRL can outperform model-free methods in sample efficiency, particularly when paired with differentiable solvers in so-called first-order gradient MBRL (FoG-MBRL) \cite{hafner2020dreamerv2, hafner2023dreamerv3, janner2019trust, clavera2020modelaugmentedactorcriticbackpropagatingpaths}.
Yet, these methods seldom leverage existing high-quality robot simulators, hence sacrificing realism, domain knowledge, and first-principle physics models. 

This work proposes to address these issues through Decoupled forward-backward Model-based policy Optimization (DMO), a first-order gradient reinforcement learning method that reduces trajectory prediction error by decoupling forward simulation (i.e., trajectory unrolling) from gradient computation. In contrast to typical MBRL approaches that rely on a single learned model for both, DMO uses a high-fidelity simulator to generate trajectories and learn the value function, while computing gradients through a differentiable learned model. This enables stable and efficient policy updates via analytical gradients, while still benefiting from GPU-accelerated simulation. Interestingly, DMO can be applied seamlessly on top of many FoG-based methods.

We demonstrate its benefits across a suite of eight control benchmarks, including dexterous manipulation, humanoid, and quadruped tasks. DMO stabilizes training, improves wall-clock time, and enables robust sim-to-real transfer, even in challenging modes such as bipedal locomotion.
DMO not only yields substantial gains in sample efficiency, by an order of magnitude over PPO, but it also consistently reduces wall-clock training time, achieving up to 20\% improvement despite the added complexity of model learning and gradient computation. It can even enable no-batch learning, a setting notably difficult \cite{elsayed2024deep}. While improved wall-clock efficiency is arguably DMO’s most immediate benefit, we believe its sample efficiency will become increasingly desirable as more complex simulators, such as foundation world models or learning directly on real robots, gain traction with significantly higher evaluation costs compared to current physics-based simulators.
Finally, policies trained with DMO are robust enough to be directly deployed on a quadruped robot.
\section{Related Work}
\paragraph{RL with model-based unrolling of trajectories}
MBRL \cite{sutton1990integrated, gu2016continuous, kurutach2018model, janner2019trust, buckman2018sample, feinberg2018model} can be used in setups where the model and the policy are trained concurrently or asynchronously. In the concurrent case, the learned dynamic model generates trajectories to train the policy, while the policy collects real samples to refine the learned model. Alternatively, the process can be asynchronous, where the dynamic model is first trained using an offline dataset and then frozen before being used to generate trajectories for policy training \cite{yu2020mopomodelbasedofflinepolicy, li2025offlineroboticworldmodel}. FoG-MBRL leverages learned dynamics models to directly backpropagate the expected sum of discounted returns through predicted trajectories \cite{Hafner2020Dream, hafner2023dreamerv3, wu2023daydreamer, ghugare2022simplifying, georgiev2024pwm, amos2021model, byravan2020imagined}. The fundamental motivation stems from the hypothesis that analytical gradient approximations provide lower-variance estimates compared to zeroth-order gradient methods \cite{clavera2020modelaugmentedactorcriticbackpropagatingpaths}, potentially enabling more sample-efficient learning. These methods construct an autodifferentiation graph from the initial state to the latest predicted state, incorporating the partial derivatives of the learned model with respect to its inputs. Then the gradient of returns can be backpropagated to the policy parameters. MAAC \cite{clavera2020modelaugmentedactorcriticbackpropagatingpaths} exemplifies this approach by unrolling trajectories using a learned model and computing policy gradients through truncated sub-trajectories, which are completed using a Q-value approximation. A notable issue of FoG-MBRL is the accumulation of prediction errors in the trajectory rollouts, which hinders training efficiency and optimality \cite{lambert2022investigatingcompoundingpredictionerrors, xiao2019learningcombatcompoundingerrormodelbased}. These models are also commonly found in planning-based approaches that use tree search methods \cite{schrittwieser2020mastering, niu2024lightzero, pu2024unizero, xuan2024rezero} or are used in conjunction with online trajectory optimization or model predictive control \cite{chua2018deep, hafner2019learning, wang2019exploring, Hansen2022tdmpc, hansen2024tdmpc2, Bechtle2019}. 



\paragraph{FoG-RL with decoupled gradient evaluation}
Prior work, such as PILCO and SVG$(\infty)$ \cite{deisenroth2011pilco, NIPS2014_6766aa27, heess2015learning}, explores computing policy gradients using derivatives of learned models along real trajectories rather than predicted ones. This method inherently avoids prediction errors accumulation.
Despite its theoretical advantages, this paradigm has been largely overlooked in subsequent research. For instance, SAC-SVG(H) \cite{amos2021model}, a successor to SVG$(\infty)$, did not include this technique, and to our knowledge, no comprehensive ablation studies have been conducted to empirically evaluate the benefits of decoupling trajectory generation and gradient computation. In this paper, we revisit this idea to enhance modern FoG-MBRL algorithms. Unlike more recent approaches, SVG$(\infty)$ optimizes cumulative returns along entire trajectories, and not along small subtrajectories (as done in e.g., MAAC \cite{clavera2020modelaugmentedactorcriticbackpropagatingpaths}), which can prove limiting. Such a decoupling has also been explored in model-free settings \cite{song2024learning} where simplified algorithms are used to approximately evaluate the simulator gradients along accurate trajectories rolled out using a high-fidelity simulator.

\paragraph{Differentiable simulators}
Differentiable simulators \cite{xu2021accelerated, georgiev2024adaptive, xing2024stabilizing} offer an alternative to learned dynamics models by directly providing analytical partial derivatives of the state transitions. 
One notable example of leveraging differentiable simulators is SHAC \cite{xu2021accelerated}, which computes policy gradients using analytical derivatives of the simulator dynamics. Another example is SAPO \cite{xing2024stabilizing}, which extends the SHAC framework by incorporating advanced policy optimization techniques such as maximum entropy regularization, state-dependent policy variance, and clipped double critic trick without target networks. These enhancements make SAPO particularly effective in exploration-heavy environments, where balancing exploration and exploitation is critical. Policies trained on Differentiable simulators have successfully been transferred to real robots \cite{bagajo2024diffsim2real, song2024learning}. Yet, their development is hindered by the challenge of efficiently computing analytical derivatives in contact-rich or multi-physics environments (e.g., soft bodies). Moreover, their non-smooth gradient landscapes may lead to inefficient optimization \cite{metz2021gradients, suh2022differentiable, lidec2022augmenting}. Learned models, on the other hand, provide by design approximate yet smooth dynamic models \cite{georgiev2024pwm} that can be used in such complex settings.
\section{Method}
We now present our main contribution: Decoupled forward-backward Model-based policy Optimization (DMO). DMO updates policy parameters using FoG estimates approximated with the derivatives of a learned model along trajectories generated by a high-fidelity simulator. By decoupling trajectory generation and gradient computation, DMO mitigates the error accumulation that often hinders FoG-MBRL algorithms, as demonstrated empirically in the next section. 

\subsection{Background}
\paragraph{Reinforcement learning}
We consider an infinite horizon, discounted Markov Decision Process (MDP) \(\left(\mathcal{S}, \mathcal{A}, r, \gamma, f \right)\), where \(\mathcal{S}\) is the state space, \(\mathcal{A}\) is the action space, \(r: \mathcal{S} \times \mathcal{A} \rightarrow \mathbb{R}\) is the reward function, \(\gamma\) is the discount factor, and \(f: \mathcal{S} \times \mathcal{A} \rightarrow \mathcal{P}(\mathcal{S})\) is the dynamics function. Here, \(\mathcal{P}(\mathcal{S})\) denotes the space of probability distributions over states. RL aims to find a policy \(\pi: \mathcal{S} \rightarrow \mathcal{P}(\mathcal{A})\) that maximizes the discounted sum of future rewards:
\begin{equation}
\max_\pi \mathbb{E}_{\tau \sim \pi, f} \left[ \sum_{t=0}^\infty \gamma^t r(s_t, a_t) \right],
\label{fo:mdp}
\end{equation}
where $\tau$ is the distribution of the trajectories under $f$ and $\pi$ and $\sum_{t=0}^\infty \gamma^t r(s_t, a_t)$ is the discounted episodic return evaluated on a particular trajectory $\tau$, denoted by $G(\tau)$. We parametrize the policy $\pi_\theta$ with a neural network.

\paragraph{Model-based RL with first order gradients}

We need the partial derivatives of the dynamics with respect to state and action to compute the gradients of the RL objective $G(\theta)$ with respect to the policy parameters. The policy gradient is estimated from the gradient of the episodic return:
\begin{align} \label{fo:apg}
    &\nabla_\theta G(\theta)=\nabla_\theta\sum^\infty_{t=0}\gamma^t r(s_t, a_t)=\sum^\infty_{t=0}\gamma^t\left [ \frac{\partial r(s,a)}{\partial s}\bigg|_{(s_t, a_t)} \frac{d s_t}{d\theta} + \frac{\partial r(s,a)}{\partial a}\bigg|_{(s_t, a_t)} \frac{d a_t}{d\theta}\right ],
\end{align}
with:
\begin{equation}
\left\{
\begin{aligned}
\frac{d s_{t+1}}{d\theta} &= \frac{\partial f(s,a)}{\partial s}\bigg|_{(s_t, a_t)}\frac{d s_t}{d\theta} + \frac{\partial f(s,a)}{\partial a}\bigg|_{(s_t, a_t)}\frac{d a_t}{d\theta} \\
\frac{d a_{t+1}}{d\theta} &= \frac{\partial \pi_\theta(s)}{\partial\theta}\bigg|_{(\tilde{\theta}, s_t)} + \frac{\partial \pi_\theta(s)}{\partial s}\bigg|_{(\tilde{\theta}, s_t)}\frac{d s_t}{d\theta}
\end{aligned}
\right.
\end{equation}


For $\frac{\partial f(s,a)}{\partial s}\bigg|_{(s_t, a_t)}$ and $\frac{\partial f(s,a)}{\partial a}\bigg|_{(s_t, a_t)}$, FoG-MBRL learns an approximation of the dynamics, $\hat{f}_\phi$, and uses it to approximate $\frac{\partial \textit{f}(s, a)}{\partial s} \approx \frac{\partial \hat{f}_\phi(s, a)}{\partial s}$ and $\frac{\partial \textit{f}(s, a)}{\partial a} \approx \frac{\partial \hat{f}_\phi(s, a)}{\partial a}$.

\subsection{Implementation of DMO}
To demonstrate the generality of our approach, we apply DMO to three distinct FoG algorithms:
\begin{enumerate}
    \item \textbf{DMO-BPTT:} This lightweight configuration uses backpropagation through time (BPTT) to compute gradients by truncating trajectories and directly propagating returns through the dynamics model. This algorithm is also known as APG \cite{freeman2021brax}. It avoids reliance on a value function, making it suitable for tasks with shorter horizons or dense rewards.
    \item \textbf{DMO-SHAC:} In this variant, truncated trajectory returns are supplemented with estimated future returns using a learned value function. This enables DMO to tackle tasks dependent on long-term or sparse rewards. It is the direct application of DMO to SHAC \cite{xu2021accelerated}.
    \item \textbf{DMO-SAPO:} This configuration incorporates SAPO's \cite{xing2024stabilizing} key enhancements to the SHAC framework, including maximum entropy regularization, state-dependent policy variance, and clipped double critic trick without target networks. This variant excels in exploration-heavy environments.
\end{enumerate}

For each implementation, we now detail how we learn the model, the critic, and the actor. Complete algorithmic details are available in Appendix \ref{appdx:algo}.
\paragraph{Model learning}
For all three versions, our proposed approach learns a model $\hat{f}_\phi$ of the dynamics by filling a replay buffer with samples encountered during policy training. $\hat{f}_\phi$ is parameterized as a multi-layer perceptron (MLP). Policy learning, value function learning (for DMO-SHAC and DMO-SAPO), and model learning happen at the same time. The outputs of $\hat{f}_\phi$ parametrize a Gaussian distribution with diagonal covariance: \[ p_{\phi}(s_{t+1} \mid s_t, a_t) = \mathcal{N}\left(\mu_{\phi}(s_t, a_t), \Sigma_{\phi}(s_t, a_t)\right). \] and is optimized using the following maximum likelihood objective:

\begin{align} \label{fo:3}
    \mathcal{L}_{\hat{f}}(\boldsymbol{\phi})=\mathbb{E}_{\left(s, a, s^{\prime}\right) \sim \mathcal{B}}\left[p_{\phi}(s^{\prime} \mid s, a)\right],
\end{align}

where $\mathcal{B}$ represents the replay buffer filled with previously observed trajectories.

\paragraph{Critic learning}
DMO-SHAC and DMO-SAPO learn an approximator $V_{\boldsymbol{\psi}}^{\pi_\theta}$, parametrized by $\psi$, to the value function of the policy derived from $\pi_\theta$: $V^{\pi_\theta}_{\boldsymbol{\psi}}(s_i)\approx \mathbb{E}_{\tau \sim \pi_\theta, f} \left[\sum_{t=i}^\infty \gamma^{t-i} r\left(s_t, a_t\right)\right]$ for DMO-SHAC and $V^{\pi_\theta}_{\boldsymbol{\psi}}(s_i)\approx \mathbb{E}_{\tau \sim \pi_\theta, f} \left[\sum_{t=i}^\infty \gamma^{t-i} \left(r\left(s_t, a_t\right)+\alpha \mathcal{H}_{\pi} \left[ a_t \mid s_t \right]\right)\right]$ for DMO-SAPO (see Appendix \ref{appdx:critic}). $\mathcal{H}_{\pi}[a_t \mid s_t]$ is the continuous Shannon entropy of the action distribution.
Here, the temperature parameter \(\alpha\) determines the trade-off between exploration (through entropy maximization) and exploitation (reward optimization). $V_{\boldsymbol{\psi}}^{\pi_\theta}$ is then used as an estimate of the value of $\pi_\theta$ to shorten the rollout horizon without adding regret when evaluating the return $G(\tau)$:
\begin{align} \label{fo:1}
    \mathcal{L}^{\text{DMO-SHAC}}_\pi(\boldsymbol{\theta}) & := \mathbb{E}_{\tau \sim \pi_\theta, f}\left[\sum_{h=1}^{H-1} \gamma^h r\left(s_h, a_h\right) + \gamma^H V^{\pi_\theta}_{\boldsymbol{\psi}}\left(s_H\right)\right],
\end{align}
\begin{align} \label{fo:5}
    \mathcal{L}^{\text{DMO-SAPO}}_\pi(\boldsymbol{\theta}) & := \mathbb{E}_{\tau \sim \pi_\theta, f}\Bigg[\sum_{h=1}^{H-1} \gamma^h 
    \big(r\left(s_h, a_h\right) + \alpha \mathcal{H}_{\pi} \left[ a_t \mid s_t \right]\big) 
      + \gamma^H V^{\pi_\theta}_{\boldsymbol{\psi}}\left(s_H\right)\Bigg].
\end{align}
Since DMO unrolls the trajectories using the high fidelity simulator, we learn the approximator $V_{\boldsymbol{\psi}}^{\pi_\theta}$ using real samples from the simulator, instead of samples generated by the learned model, as is usually done in MBRL. DMO-BPTT truncates the horizon and simply uses the gradient of the following loss:
\begin{align} \label{fo:4}
    \mathcal{L}^{\text{DMO-BPTT}}_\pi(\boldsymbol{\theta}):=\mathbb{E}_{\tau \sim \pi_\theta, f}\left[\sum_{h=1}^{H-1}  r\left(s_h, a_h\right)\right].
\end{align}



\paragraph{Actor learning}
The actor $\pi_\theta$ is parametrized as an MLP. Its outputs parametrize a Gaussian distribution with diagonal covariance $ \pi_{\theta}(a_t \mid s_t) = \mathcal{N}\left(\mu_{\theta}(s_t), \Sigma_{\theta}\right)$, for DMO-SHAC and DMO-BPTT, and $\pi_{\theta}(a_t \mid s_t) = \mathcal{N}\left(\mu_{\theta}(s_t), \Sigma_{\theta}(s_t)\right)$ for DMO-SAPO. The actor is learned through gradient descent on the shortened returns on a batch of rollouts. During the forward pass of the optimization algorithm, the simulator is used to get the true next state $s_{t+1}=f(s_t,a_t)$, unlike previous FoG-MBRL methods that use $\hat{f}_\phi$ to predict $\hat{f}_\phi(\hat{s}_t,a_t)=\hat{s}_{t+1}\approx s_{t+1}$. During the backward pass, however, $\hat{f}_\phi$ is used to approximate the partial derivatives of the true dynamics $f$. But instead of using the partial derivatives of $\hat{f}_\phi$ taken at $\hat{s}_{t+1}$, $\frac{\partial \hat{f}_\phi(s, a)}{\partial s} \bigg|_{(\hat{s}_{t+1},a_{t+1})}$, DMO uses the partial derivatives of $\hat{f}_\phi$ taken at $s_{t+1}$, $\frac{\partial \hat{f}_\phi(s, a)}{\partial s} \bigg|_{(s_{t+1},a_{t+1})}$. This separation—using the simulator for forward trajectory unrolling and the learned model for gradient computation—represents what we refer to as "decoupling." Unlike most approaches, where both forward and backward passes are performed using the same function (e.g., either the simulator or the learned model), DMO explicitly separates these two processes by using different functions. We use a stochastic policy $\pi_\theta$ for exploration purposes. We use the reparametrization trick \citep{kingma2022autoencodingvariationalbayes} to sample from $\pi_\theta$ and compute a valid gradient with it (see Appendix \ref{appdx:reparam_trick}). We rely on the automatic differentiation framework of PyTorch \citep{NEURIPS2019_bdbca288} to compute the complete first-order gradient estimate of (\ref{fo:1}) (DMO-SHAC), (\ref{fo:5}) (DMO-SAPO), or (\ref{fo:4}) (DMO-BPTT). 
Additionally, we propose an efficient numerical implementation that removes the need for computation graph manipulation (detailed in Appendix \ref{appdx:pytorch}) and allows the integration of decoupling in very few lines of code.
\section{Experiments}
\subsection{Experimental Setup}
\paragraph{Simulation Environments}
\begin{figure}[t]
    \centering
    \begin{minipage}{1.0\textwidth}
        \centering
        \begin{subfigure}{0.16\textwidth}
            \centering
            \text{Ant}\\[0.2ex]
            \includegraphics[width=\textwidth]{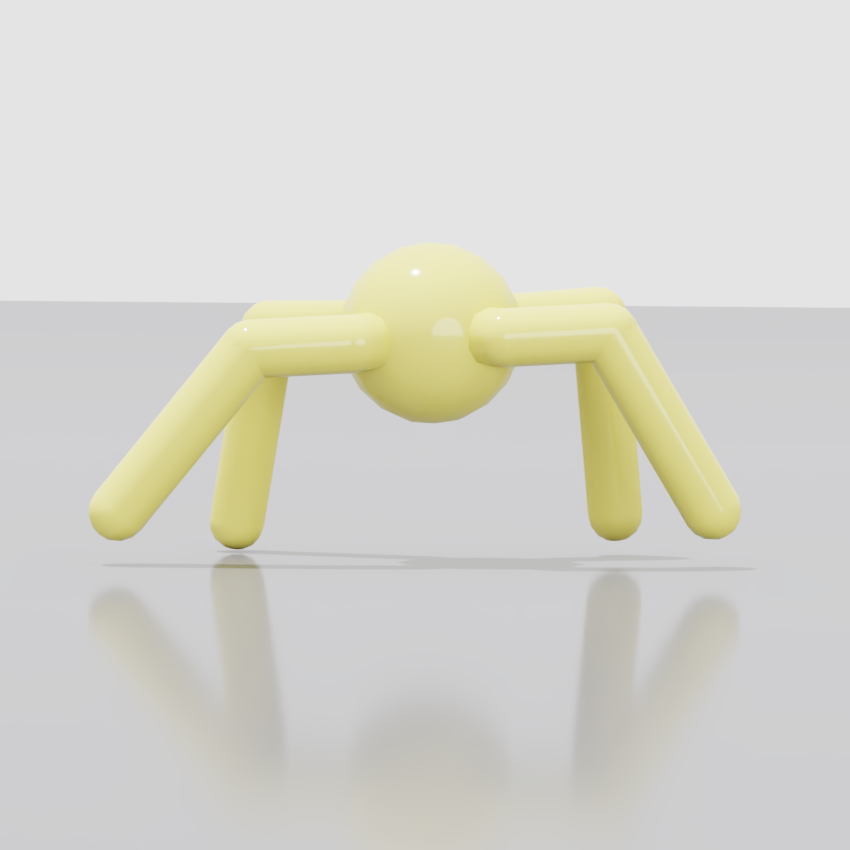}
        \end{subfigure}
        \begin{subfigure}{0.16\textwidth}
            \centering
            \text{SNU Humanoid}\\[0.2ex]
            \includegraphics[width=\textwidth]{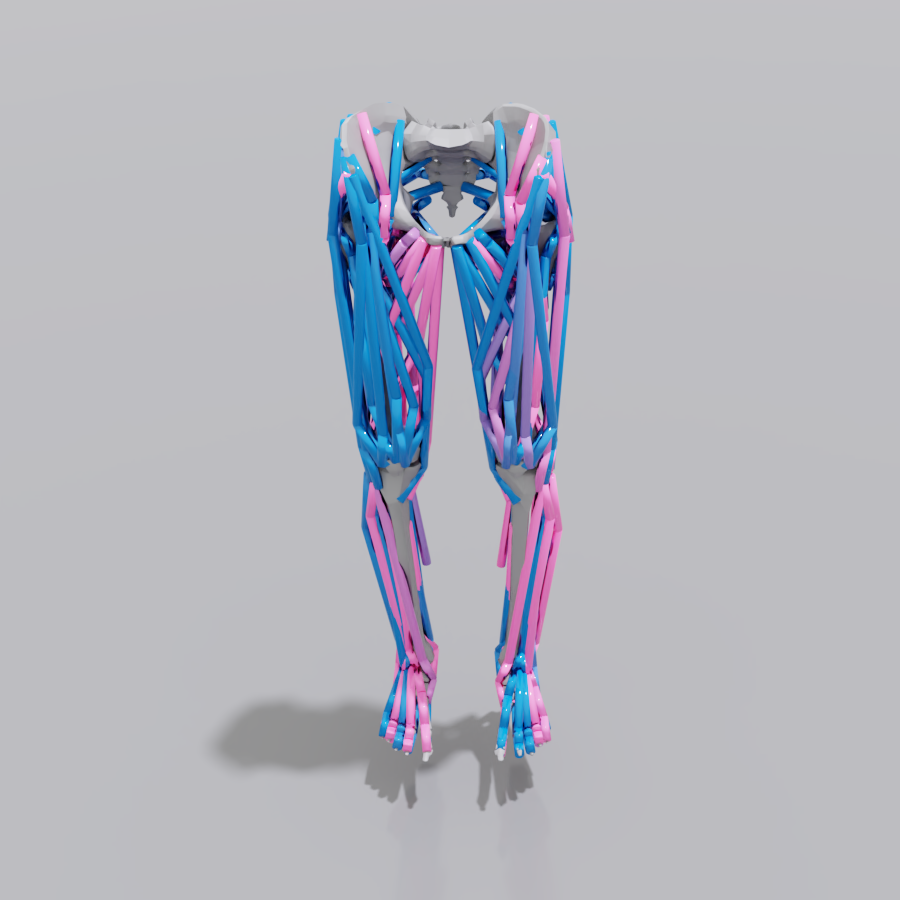}
        \end{subfigure}
        \begin{subfigure}{0.16\textwidth}
            \centering
            \text{Cheetah}\\[0.2ex]
            \includegraphics[width=\textwidth]{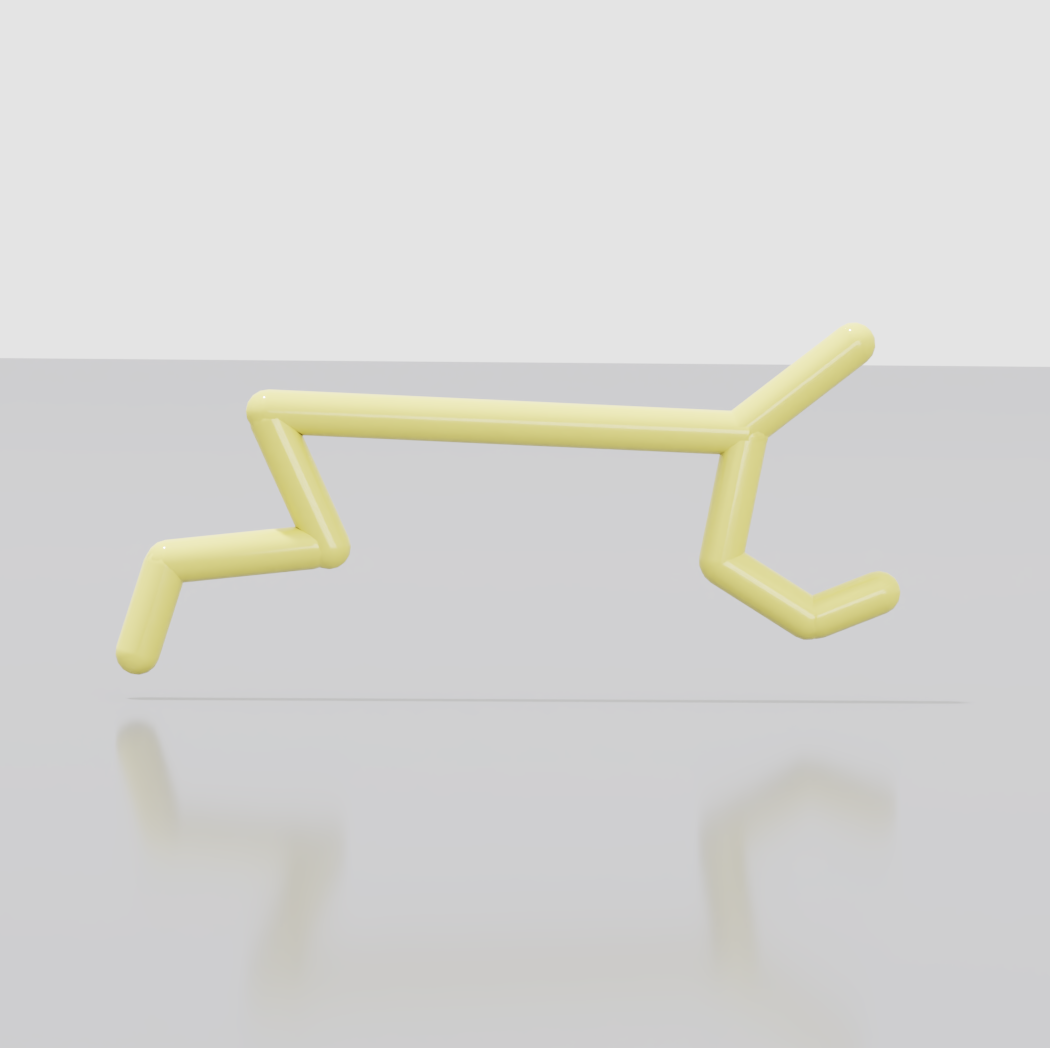}
        \end{subfigure}
        \begin{subfigure}{0.16\textwidth}
            \centering
            \text{Hopper}\\[0.2ex]
            \includegraphics[width=\textwidth]{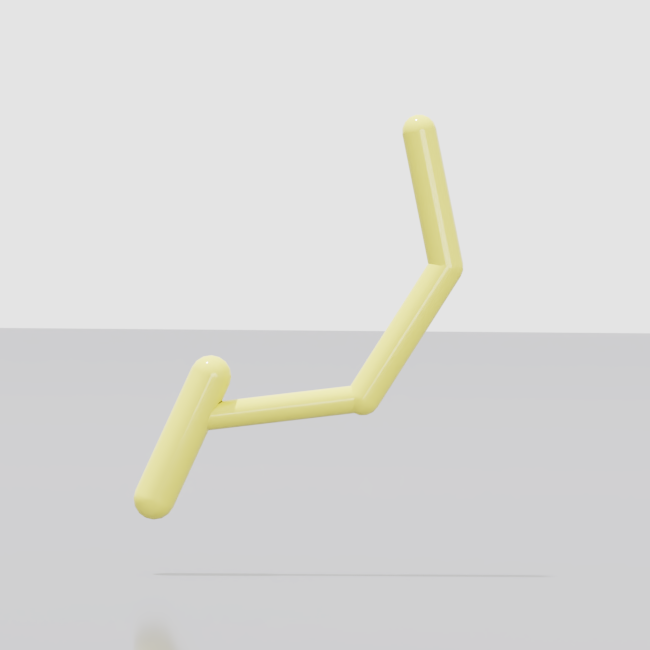}
        \end{subfigure}
        \begin{subfigure}{0.16\textwidth}
            \centering
            \text{Allegro Hand}\\[0.2ex]
            \includegraphics[width=\textwidth]{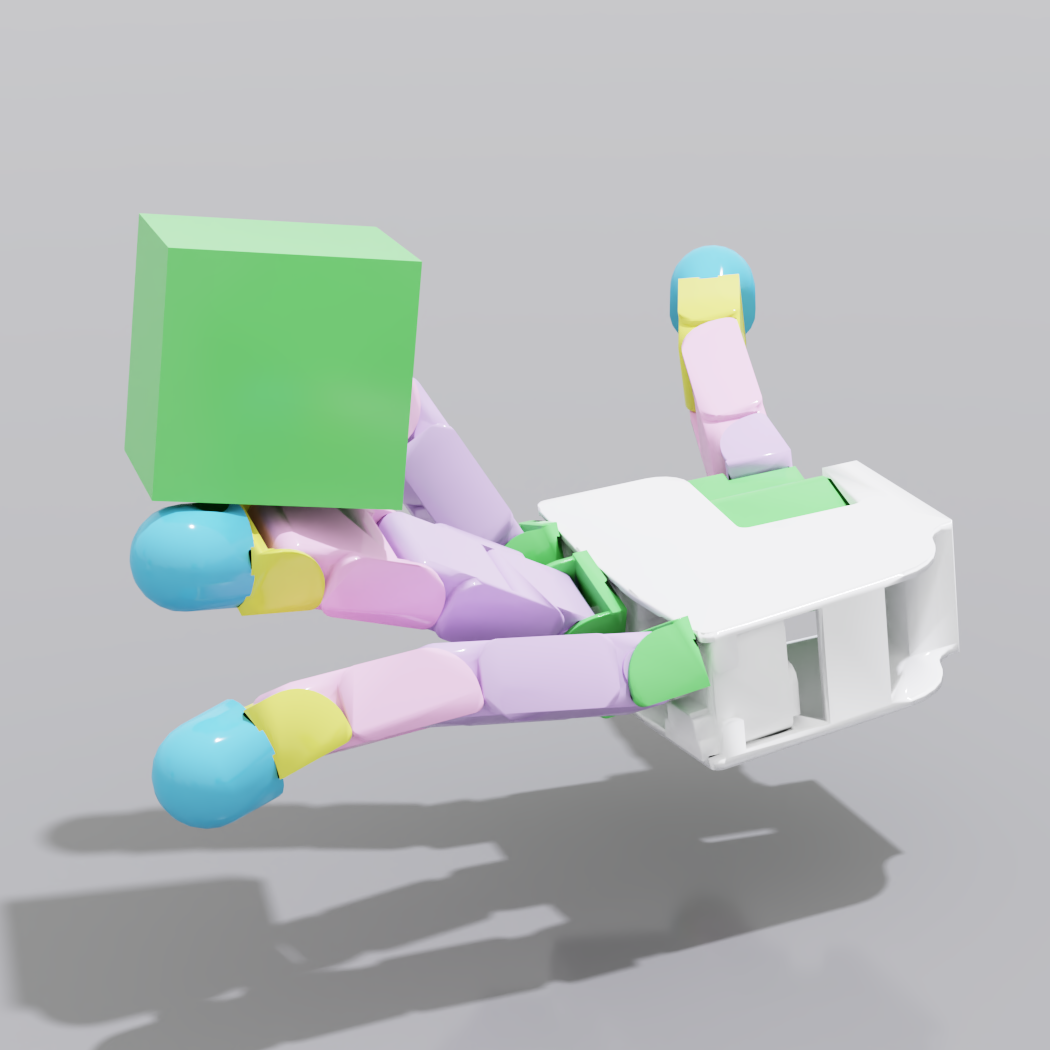}
        \end{subfigure}
        \begin{subfigure}{0.16\textwidth}
            \centering
            \text{Humanoid}\\[0.2ex]
            \includegraphics[width=\textwidth]{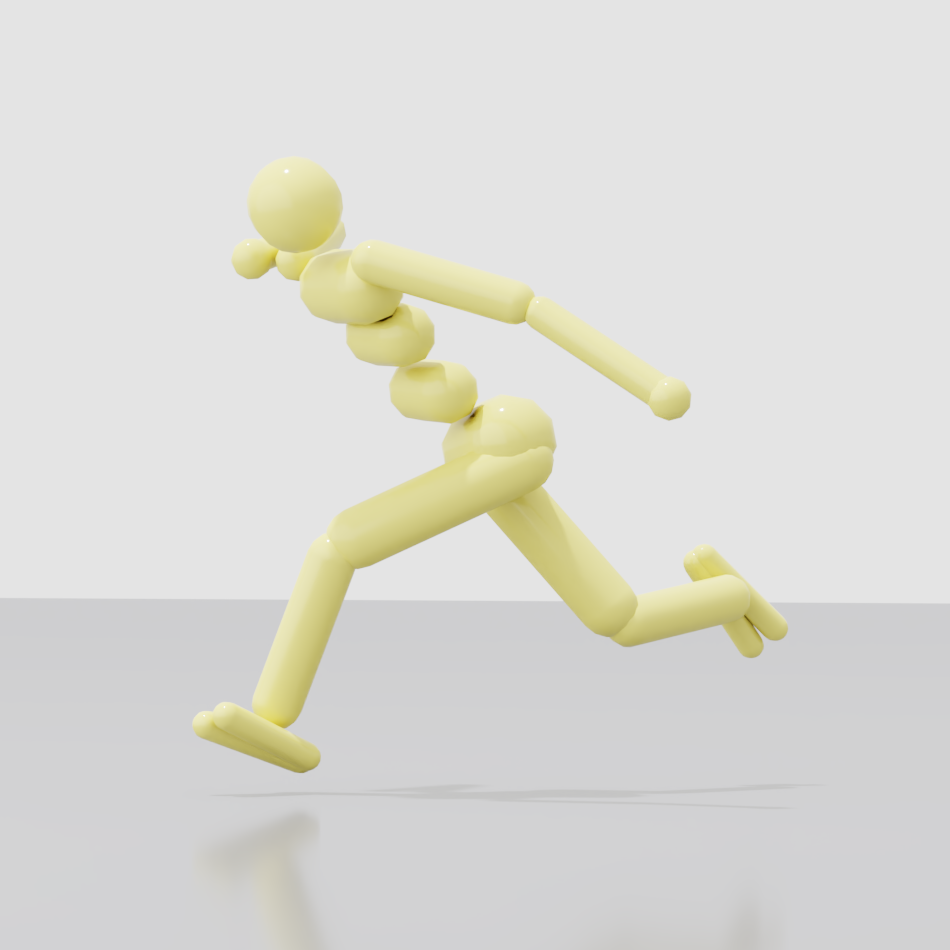}
        \end{subfigure}
    \end{minipage}
    \caption{\textbf{Visualizations of Environments Trained with DMO.} Each image represents a distinct simulation environment: Ant, SNU Humanoid, Cheetah, Hopper, Allegro Hand, and Humanoid.}
    \label{fig:env_visualizations}
\end{figure}

We conduct most of our experiments on the environments provided by the DFlex simulator \cite{xu2021accelerated} (Figure \ref{fig:env_visualizations}). DFlex is a GPU-accelerated differentiable simulator. It comes with six already implemented environments: Ant, Hopper, Cheetah, Humanoid, and SNUHumanoid. We chose this simulator as it satisfied two key requirements for our method: (1) DFlex parallelizes the simulation on the GPU, and (2) the reward functions of the provided environments are designed for and thus compatible with FoG methods. For these environments, we only report results of DMO-SHAC, as DMO-BPTT did not perform well, and DMO-SAPO achieved comparable results on them. We added to this benchmark the AllegroHand environment adapted for FoG methods \cite{xing2024stabilizing}. 

\paragraph{Real Robot Experiment Setup for Quadrupedal Motion} We trained a velocity-commanded walking policy for the Unitree Go2 quadruped robots using DMO and successfully deployed it on the real robot. For the simulation, we used the GPU-accelerated non-differentiable IsaacGym simulator \cite{makoviychuk2021isaac}. To facilitate sim-to-real transfer, we incorporated a simple joint friction model during training and randomized its parameters \cite{duclusaud2025extendedfrictionmodelsphysics}. The reward function used for training was adapted from \citet{margolis2023walk}, with modifications to fix the gait parameters that were originally designed as adjustable commands. A detailed description of the reward and the observations is given in Appendix \ref{appdx:go2_quadrupedal_reward}. The policy sends desired positions, which are tracked with a low-level PD controller.

\paragraph{Real Robot Experiment Setup for Bipedal Motion with Go2} To demonstrate the ability of DMO to generate dynamic behaviors, we developed a bipedal locomotion environment for the Go2 quadruped robot in IsaacGym, where the robot must transition from standing on all four legs to steady balancing on its front legs. The reward and early stopping structure, inspired by \cite{li2024learning}, encourages both the initial lifting motion and sustained balance. This formulation requires sufficient exploration to discover effective strategies for transitioning to and maintaining a bipedal stance. As expected, DMO-SHAC and DMO-BPTT did not perform well due to their poor exploration abilities, so we only report results for DMO-SAPO.

\paragraph{Baselines for Comparison}
We evaluate our algorithm against PPO \cite{schulman2017proximalpolicyoptimizationalgorithms} and SAC \cite{haarnoja2018soft, haarnoja2019softactorcriticalgorithmsapplications}, two model-free RL algorithms widely used in robotics. We also compare against MAAC \cite{clavera2020modelaugmentedactorcriticbackpropagatingpaths}, a FoG-MBRL method. For MAAC, we implemented a modernized version using parallel data collection, gathering a batch of samples at each environment step instead of a single sample. This modification significantly improved the wall-clock efficiency of the algorithm. Additionally, we employed the same value function learning scheme as DMO-SHAC, when compared to DMO-SHAC, or DMO-SAPO, when compared to DMO-SAPO, as it generally outperforms standard TD-learning and enables a fairer comparison. Note that the hyperparameters for the original MAAC algorithm were not publicly available.

\subsection{Results and Analysis}
\paragraph{Sample and Time Efficiency}
\begin{figure}[h!]
    \centering
    \begin{minipage}{0.49\linewidth}
        \centering
        \includegraphics[width=\linewidth]{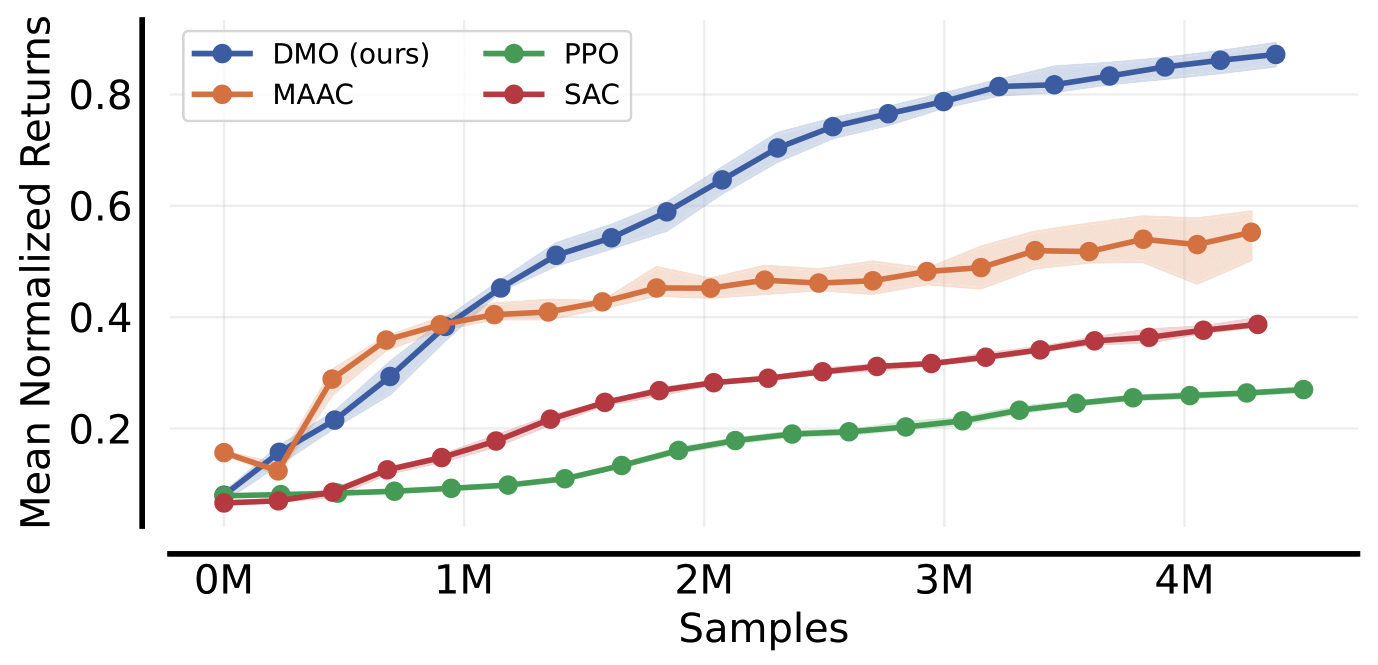}
    \end{minipage}
    \hfill
    \begin{minipage}{0.49\linewidth}
        \centering
        \includegraphics[width=\linewidth]{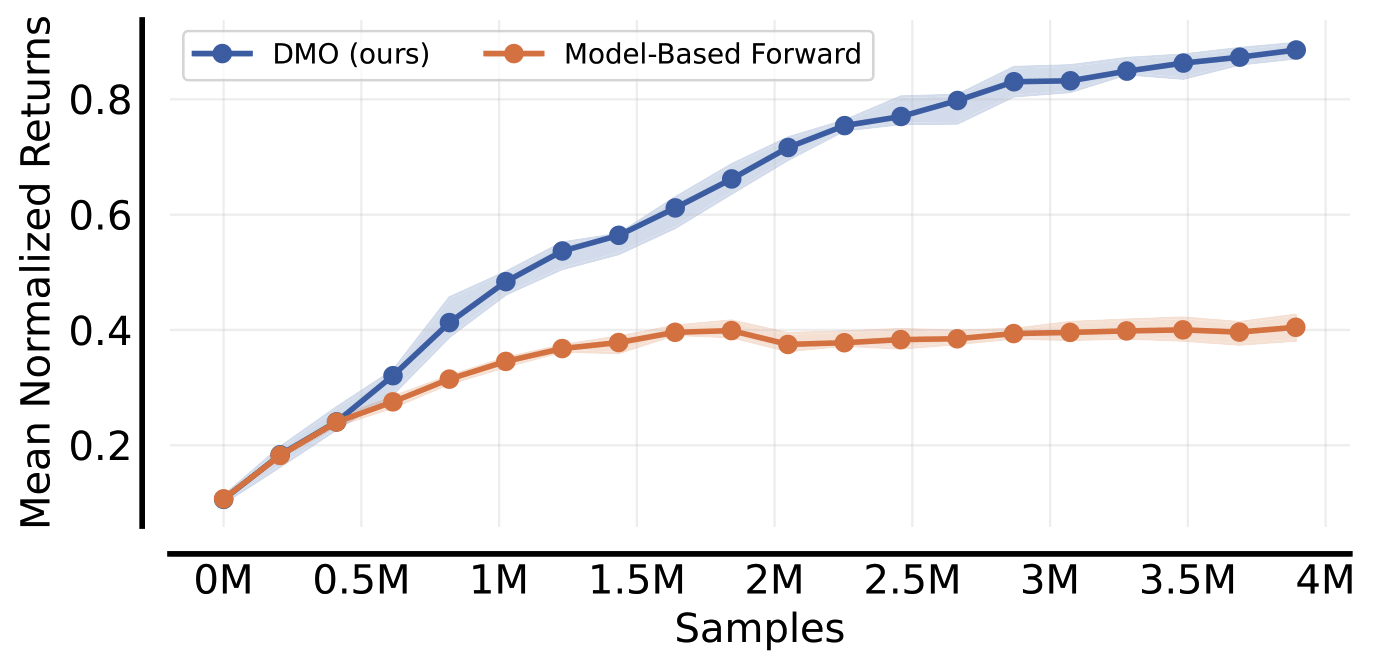}
    \end{minipage}
    \caption{
        \textbf{Left: Sample Efficiency at 4M Samples.} Results for DMO, PPO, SAC, and MAAC, all limited to 4M samples. \textbf{Right: Sample Efficiency with Model-Based Ablation.} Comparison of DMO to its counterpart that uses learned model forward passes, both at 4M samples. Aggregate normalized scores with mean and 95\% confidence intervals over all environments and 5 seeds are shown.
    }
    \label{fig:combined_efficiency}
\end{figure}

\begin{figure}[h!]
    \centering
    \begin{minipage}{0.49\linewidth}
        \centering
        \includegraphics[width=\linewidth]{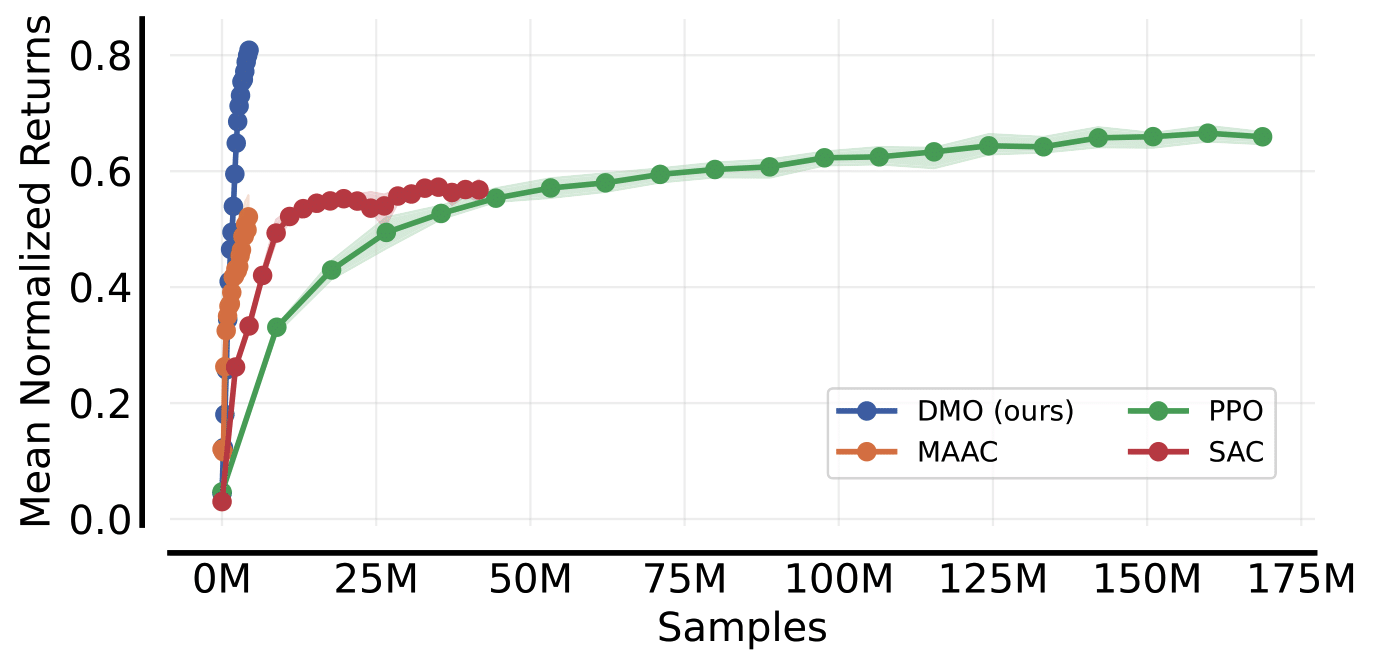}
    \end{minipage}
    \hfill
    \begin{minipage}{0.49\linewidth}
        \centering
        \includegraphics[width=\linewidth]{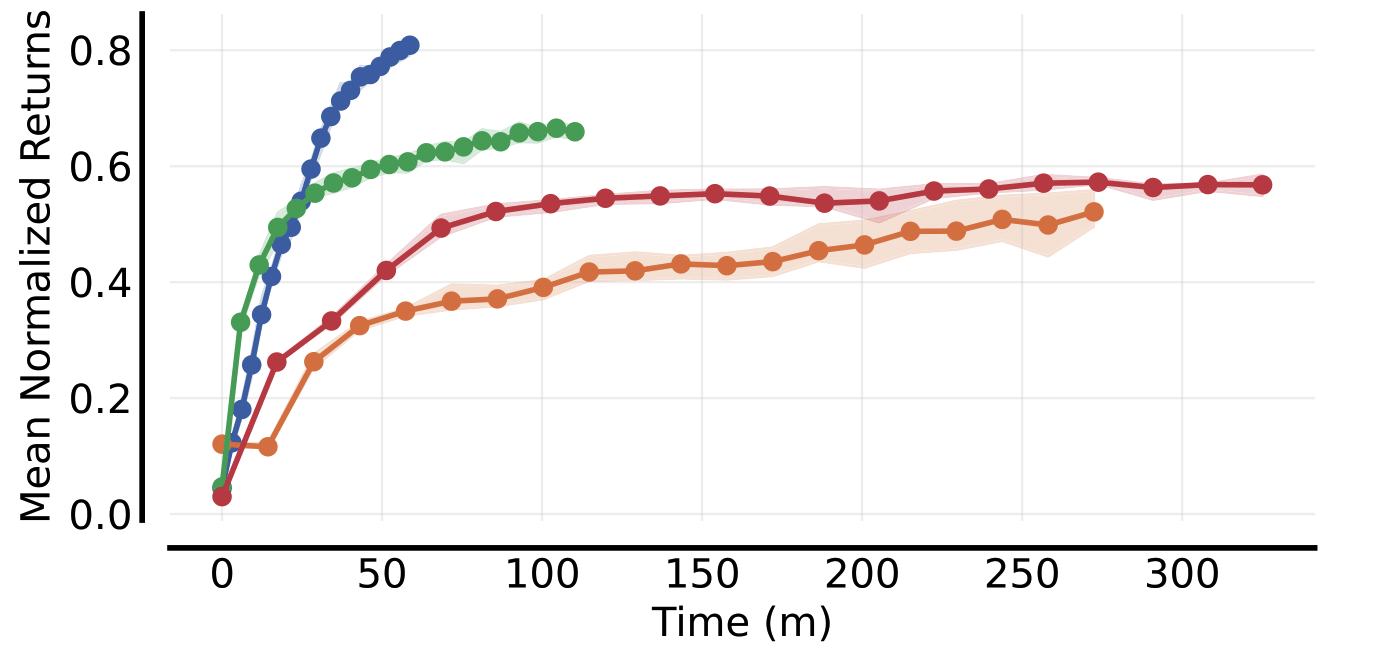}
    \end{minipage}
    \caption{
        \textbf{Left: Sample Efficiency at Extended Training.} Results for DMO and MAAC using 4M samples, while PPO is trained for 160M samples and SAC for 40M samples. \textbf{Right: Wall-Clock Time Efficiency at Extended Training.} Results for the same sample allocations as on the left. Aggregate normalized scores with mean and 95\% confidence intervals across all environments and 5 seeds are shown.
    }
    \label{fig:extended_efficiency_comparison}
\end{figure}


In our experiments, DMO achieves asymptotic convergence with fewer than 4 million samples, over ten times less than PPO, highlighting the efficiency of our method. Figure~\ref{fig:combined_efficiency} (left) showcases DMO dominance in sample efficiency across algorithms. Furthermore, as shown in Figure~\ref{fig:extended_efficiency_comparison} (left), even when PPO is trained with 160M samples and SAC with 40M samples---while DMO and MAAC remain at 4M samples---DMO maintains unparalleled sample efficiency and surpasses the asymptotic performance of the compared methods, including PPO, demonstrating its potential for rapid, resource-efficient training. Individual sample efficiency curves for every environment are available in Appendix \ref{appdx:detailed_results}. Figure~\ref{fig:extended_efficiency_comparison} (right) further illustrates DMO superior wall-clock time efficiency across all environments compared to PPO (160M samples) and SAC (40M samples), with DMO and MAAC still using only 4M samples. This result is particularly noteworthy, as prior model-based RL methods like MAAC often suffer from significant computational overhead---evident in MAAC prolonged training time for just 4M samples---negating their theoretical sample efficiency gains. DMO, in contrast, delivers both sample and time efficiency. We also compared with the main baselines in first-order RL (SHAC/SAPO) and have shown that DMO is competitive in both efficiency and final convergence accuracy, despite it is not using the extra information available to these algorithms (see Appendix \ref{appdx:detailed_results}).

\begin{figure}[t]
    \centering
    \includegraphics[width=0.49\linewidth]{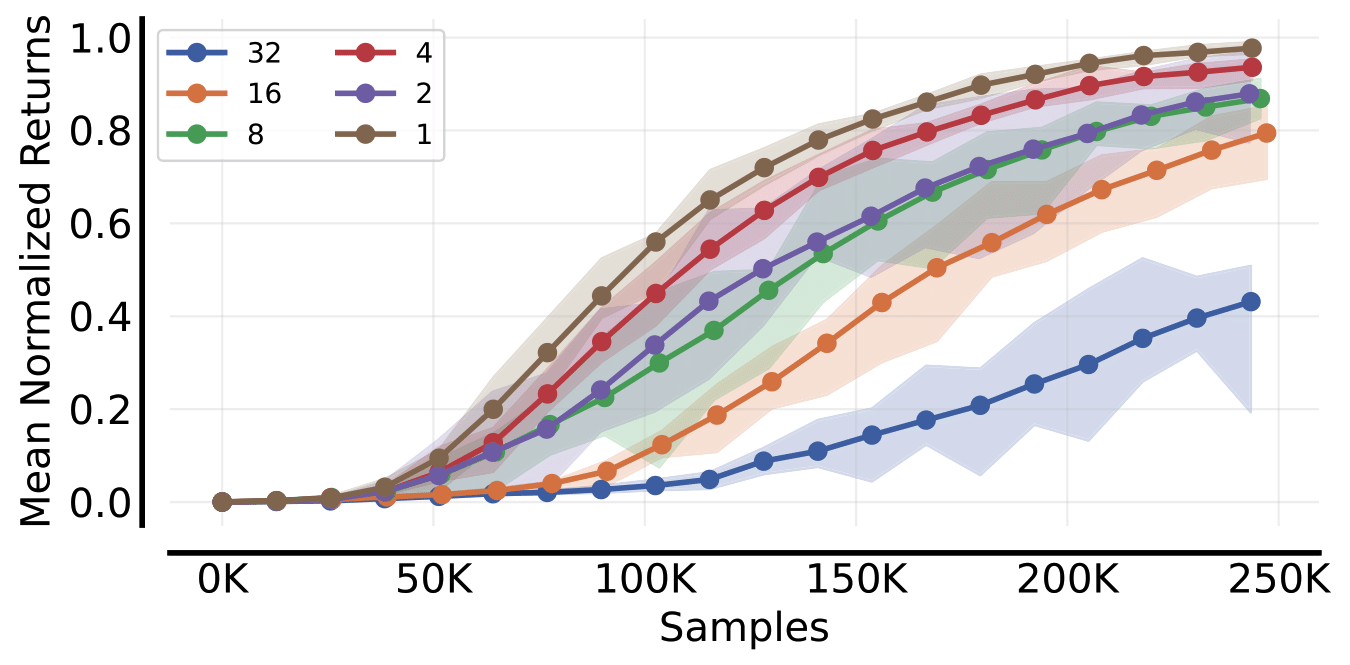}
    \includegraphics[width=0.49\linewidth]{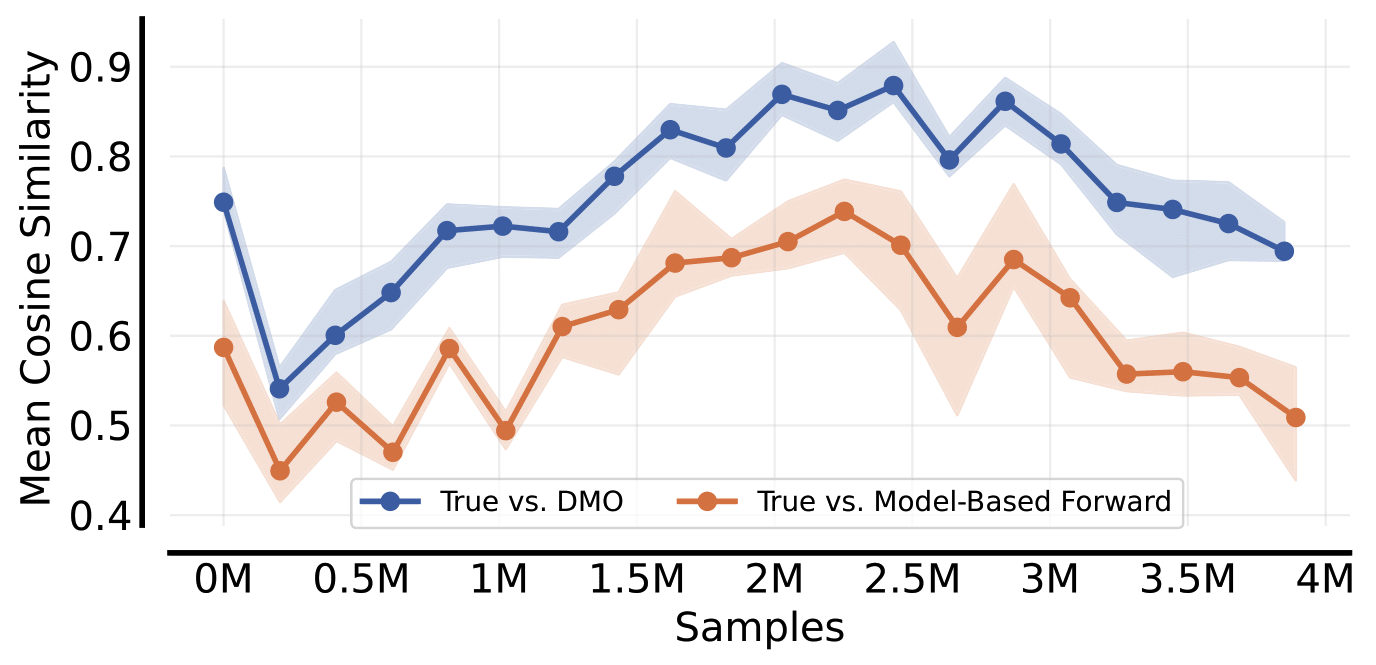}
    \caption{
        \textbf{Left: Sample Efficiency of DMO-BPTT Across Batch Sizes for the Go2 Quadrupedal Task.} This illustrates the sample efficiency of DMO-BPTT for various batch sizes, with mean and 95\% confidence intervals calculated over 5 seeds. Notably, sample efficiency increases as batch size decreases to 1, and DMO-BPTT remains stable even in this configuration.
        \textbf{Right: Cosine Similarity of Gradient Computations.} This shows the cosine similarity (higher values indicate greater similarity) between gradients computed with DFlex and DMO, and between gradients computed with DFlex and the DMO counterpart that uses learned model forward passes. Results are aggregated across all five DFlex environments, with mean and 95\% confidence intervals calculated over 5 seeds.
    }
    \label{fig:combined_go2_cosine_efficiency}
\end{figure}

\paragraph{Go2 Walking Environment Analysis}
In the Go2 environment, we made two key discoveries. First, the considered reward function inspired from \citet{margolis2023walk} is densely informative, hence it eliminates the need for the value function in (\ref{fo:1})-(\ref{fo:5}), enabling the use of DMO-BPTT, similar to \citep{song2024learning}. Indeed, the considered reward ensures that the optimal behavior over a short horizon of 16 steps (like the one employed in this study) is close to the optimal behavior over an infinite horizon. Second, removing the value function allows training without batch updates, resulting in the most sample-efficient configuration. As shown in Figure \ref{fig:combined_go2_cosine_efficiency} (left), sample efficiency increases as the batch size of policy updates decreases, reaching its top efficiency at a batch size of 1. This finding is surprising, as very small batch sizes typically degrade gradient estimate quality.

\paragraph{Real Robot Deployment}
Experiments on the real robot demonstrate that the learned policies, for both the quadrupedal and bipedal tasks, are robust and transferable. A video of the policies deployed on the real robot is available in the supplementary material.

\subsection{Ablative analysis}
\paragraph{Ablation Study on Decoupling Effect}
To isolate the impact of decoupling trajectory simulation from gradient computation in DMO, we conducted an ablation study by modifying our algorithm to use the learned model for forward passes, akin to traditional model-based RL (MBRL) approaches. In this variant, termed "Model-Based Forward," the gradients $\frac{\partial \hat{f}_\phi(s, a)}{\partial s} \bigg|_{(s_{t+1}, a_{t+1})} \quad \text{and} \quad \frac{\partial \hat{f}_\phi(s, a)}{\partial a} \bigg|_{(s_{t+1}, a_{t+1})}$ are replaced by approximations computed from the learned model \textit{along trajectories predicted by the model},
$\frac{\partial \hat{f}_\phi(s, a)}{\partial s} \bigg|_{(\hat{s}_{t+1}, a_{t+1})} \quad \text{and} \quad \frac{\partial \hat{f}_\phi(s, a)}{\partial a} \bigg|_{(\hat{s}_{t+1}, a_{t+1})}.$ As shown in Figure~\ref{fig:combined_efficiency} (right), the comparison between DMO and "Model-Based Forward" reveals the substantial benefit of decoupling. With the only difference being the use of decoupled gradients, DMO achieves nearly double the asymptotic performance on average, underscoring the critical role of this design choice in enhancing sample efficiency and overall effectiveness.



\paragraph{Ablative Analysis of Gradients} We conducted an experiment where we unrolled three trajectories in parallel, constructing separate backpropagation graphs for each: (1) DFlex trajectories, where partial derivatives of the dynamics were provided by the differentiable simulator; (2) DMO-SHAC trajectories, based on our method; and (3) trajectories unrolled with the learned model, generated under the exact same conditions as DMO-SHAC. For each trajectory, we computed the policy gradient and measured the cosine similarity between the DFlex gradient and the DMO gradient, as well as between the DFlex gradient and the gradient of the model-predicted trajectory. The DMO gradients were used to update the policy. Figure \ref{fig:combined_go2_cosine_efficiency} (right) presents these cosine similarities, demonstrating empirically that decoupled gradients provide more precise policy updates.
\section{Conclusion}
We introduced Decoupled forward-backward Model-based policy Optimization (DMO), an approach that leverages GPU-accelerated simulators and decoupled gradient computation to achieve state-of-the-art sample and wall-clock efficiency in reinforcement learning for robotics. By separating trajectory generation from gradient estimation, DMO significantly stabilizes learning and mitigates error accumulation compared to prior FoG-MBRL methods. Notably, DMO significantly outperforms PPO, converging with tenfold fewer samples, exceeding its asymptotic performance, and showing better time efficiency. Our experiments across diverse environments, including real-world deployment on the Go2 quadruped robot for both walking and bipedal tasks, demonstrate its practical effectiveness and transferability. Future work could explore whether FoG-MBRL methods can scale effectively and remain competitive on even more complex real-world tasks, such as performing Parkour using first-person depth camera inputs, which are currently dominated by other RL approaches.


\clearpage
\section*{Limitations}
Our proposed method has two primary limitations. First, it requires differentiable reward functions. Many existing reward designs incorporate discrete components, such as survival bonuses, which produce zero gradients. In such scenarios, learning heavily depends on the value function, undermining the structural benefits of First-Order Gradient (FoG) methods. Consequently, reward functions often need to be redesigned to ensure compatibility, which can be a non-trivial task.

Second, our approach employs a simplistic world model, specifically an MLP regressing the next state given a state-action input. This model is ill-suited for complex inputs like images or point clouds, where more sophisticated models, such as those in \citep{hansen2024tdmpc2, li2025offlineroboticworldmodel}, would be more appropriate. However, this limitation is orthogonal to our core contribution of decoupled gradient computation, and integrating advanced world models with our method remains a promising direction for future work.



\acknowledgments{We used Open RL Benchmark \cite{Huang_Open_RL_Benchmark_2024} to generate the graphs shown in this paper. This work was in part supported by the National Science Foundation grants 1932187, 2026479, 2222815 and 2315396, the French National Research Agency Artificial and Natural Intelligence Toulouse Institute (ANR 19-P3IA-0004), the AGIMUS project, funded by the European Union under GA no.101070165, the Dynamograde joint laboratory (grant ANR-21-LCV3-0002), and the ANR NERL project (grant ANR-23-CE94-0004). It was granted access to the HPC resources of IDRIS under the allocations AD011015316R1 made by GENCI.}


\bibliography{references}  

\appendix
\appendix
\section{Appendix}

\subsection{Algorithmic Details}
\subsubsection{Critic learning} \label{appdx:critic}
$V^{\pi_\theta}_{\boldsymbol{\psi}}$ is learned with SGD by optimizing:
\begin{align}
    \mathcal{L}_V(\boldsymbol{\psi}):=\sum_{h=1}^{H-1}\left\|V^{\pi_\theta}_{\boldsymbol{\psi}}\left(s_h\right)-\hat{V}\left(s_h\right)\right\|_2^2, \label{fo:2}
\end{align}
where:
\begin{align}
    V_h\left(s_t\right):=\sum_{n=t}^{t+h-1} \gamma^{n-t} r\left(s_n, a_n\right)+\gamma^{t+h} V^{\pi_\theta}_{\boldsymbol{\psi}}\left(s_{t+h}\right) \\
    \hat{V}\left(s_t\right):=(1-\lambda)\left[\sum_{h=1}^{H-t-1} \lambda^{h-1} V_h\left(s_t\right)\right]+\lambda^{H-t-1} V_H\left(s_t\right)
 \end{align}
\subsubsection{DMO algorithm} \label{appdx:algo}
The complete algorithm is detailed in Algorithm \ref{alg:main}. The differences between DMO and SHAC only are colored in \textcolor{red}{red}, the differences between DMO and MAAC only are colored in \textcolor{blue}{blue}, while the differences with DMO that appear in both SHAC and MAAC are colored in \textcolor{orange}{orange}.
\begin{algorithm}
\caption{DMO algorithm} \label{alg:main}
\begin{algorithmic} 
\For{epoch = 1 to N}
    \State \textcolor{red}{\# Dynamical model learning}
    \For{\textcolor{red}{model mini epoch}}
        \State\textcolor{red}{$\left(s, a, s^{\prime}\right) \sim \mathcal{B}$}
        \State\textcolor{red}{$\boldsymbol{\phi} \leftarrow \boldsymbol{\phi}+\alpha_{\boldsymbol{\phi}} \nabla_\phi\mathcal{L}_{\hat{f}}(\boldsymbol{\phi})$}
    \EndFor
    \State \# Actor and critic learning
    \State $total\_reward \gets 0$
    \For{h = 1 to H}
        \State $a_h \gets \pi_\theta(s_h)$
        \State \textcolor{blue}{$s_{h+1} \gets f(s_h, a_h)$ (instead of $\hat{s}_{h+1}\gets f(\hat{s}_h, a_h)$)}
        \State $r_h \gets r(s_{h+1}, a_h)$
        \State $total\_reward \gets total\_reward - r_h$
    \EndFor
    \State Either compute $\mathcal{L}_\pi(\boldsymbol{\theta})=\mathcal{L}^{\text{DMO-SHAC}}_\pi(\boldsymbol{\theta})/\mathcal{L}^{\text{DMO-SAPO}}_\pi(\boldsymbol{\theta})$ from $total\_reward$ and the value $V_{\boldsymbol{\psi}}^{\pi_\theta}(s_{H+1})$ or $\mathcal{L}_\pi(\boldsymbol{\theta})=\mathcal{L}^{\text{DMO-BPTT}}_\pi(\boldsymbol{\theta})$ from $total\_reward$
    \State \textcolor{blue}{Compute $\mathcal{L}_V(\boldsymbol{\psi})$ from $(s_1, \dots, s_H)$ (instead of $(\hat{s}_1, \dots, \hat{s}_H)$)}
    \State \textcolor{orange}{Use $\frac{\partial \hat{f}_\phi(s, a)}{\partial s} \bigg|_{(s_{t+1},a_{t+1})}$ and $\frac{\partial \hat{f}_\phi(s, a)}{\partial a} \bigg|_{(s_{t+1},a_{t+1})}$ to approximate $\frac{\partial f(s, a)}{\partial s} \bigg|_{(s_{t+1},a_{t+1})}$ and $\frac{\partial f(s, a)}{\partial a} \bigg|_{(s_{t+1},a_{t+1})}$ during the following backward pass.}
    \State $\nabla_\theta\mathcal{L}_\pi(\boldsymbol{\theta}) \gets \text{backward}(\mathcal{L}_\pi(\boldsymbol{\theta}))$
    
    \State $\boldsymbol{\theta} \leftarrow \boldsymbol{\theta}+\alpha_{\boldsymbol{\theta}} \nabla_\theta\mathcal{L}_\pi(\boldsymbol{\theta})$
    \State $\boldsymbol{\psi} \leftarrow \boldsymbol{\psi}+\alpha_{\boldsymbol{\psi}} \nabla_\psi\mathcal{L}_V(\boldsymbol{\psi})$ (for DMO-SHAC/SAPO only)
\EndFor
\end{algorithmic}
\end{algorithm}

\clearpage
\subsubsection{PyTorch Decoupling Implementation} \label{appdx:pytorch}
\begin{lstlisting}[style=python, caption={\textbf{Gradient Swapping Function in PyTorch}}, label={code:gradient_swapping}]
class GradientSwapingFunction(Function):
    @staticmethod
    def forward(ctx, img_next_state, real_next_state):
        return real_next_state.clone()

    @staticmethod
    def backward(ctx, grad_real_next_state):
        return grad_real_next_state, None

with torch.no_grad():
    real_next_obs_, rew, done, extra_info = env.step(actions)
img_next_obs = dyn_model(obs, actions)
real_next_obs = GradientSwapingFunction.apply(img_next_obs, real_next_obs_.clone())
\end{lstlisting}

Code \ref{code:gradient_swapping} illustrates how decoupling can be implemented using PyTorch's automatic differentiation framework with minimal code and without the need for complex manual modifications to the backpropagation graph. The key idea is to rely on the simulator to compute \(s_{t+1}\) from \(s_t\) and \(a_t\), while simultaneously using the learned dynamics model to compute \(\hat{s}_{t+1}\) from the same inputs. The \verb|GradientSwapingFunction| class ensures that the backpropagation flows through \(\hat{s}_{t+1}\), as required for gradient computations while keeping \(s_{t+1}\) as a leaf node.

The implementation proceeds as follows. The simulator stepping function computes the real next state \(s_{t+1}\) given \(s_t\) and \(a_t\), while the dynamics model predicts \(\hat{s}_{t+1}\) from \(s_t\) and \(a_t\). The \verb|forward| method of the \verb|GradientSwapingFunction| class takes both \(\hat{s}_{t+1}\) and \(s_{t+1}\) as inputs. It returns a copy of \(s_{t+1}\), denoted as \(\bar{s}_{t+1}\), which is inserted into the backpropagation graph. Then the \verb|backward| method ensures that, during backpropagation, PyTorch flows the gradient \(\frac{\partial \mathcal{L}_\pi(\boldsymbol{\theta})}{\partial \bar{s}_{t+1}}\) (from the objective function in Equation \ref{fo:1}) back to \(\hat{s}_{t+1}\). Finally, since \(\hat{s}_{t+1}\) is predicted from $s_t$ and $a_t$, the gradient further flows back to $s_t$ and $a_t$. This is repeated recursively and produces the correct decoupled gradient approximation.

Formally, the backpropagation step taken in these settings from $\bar{s}_{t+1}$ to $s_t$ and $a_t$ can be written as:
\[
\frac{d \hat{s}_{t+1}}{d s_t} \frac{d\mathcal{L}_\pi(\boldsymbol{\theta})}{d \bar{s}_{t+1}} \text{ and } \frac{d \hat{s}_{t+1}}{d a_t} \frac{d\mathcal{L}_\pi(\boldsymbol{\theta})}{d \bar{s}_{t+1}}.
\]
Since \(\bar{s}_{t+1}\) is a copy of \(s_{t+1}\), and $\hat{s}_{t+1}=\hat{f}(a_t,s_t)$, this can be rewritten as:
\[
\frac{\partial \hat{f}(s, a)}{\partial s}\bigg|_{(s_t, a_t)} \frac{d\mathcal{L}_\pi(\boldsymbol{\theta})}{d s_{t+1}}.
\]
This is the decoupling formula. By design, \(\hat{f}(s, a)\) is trained to match the simulator \(f(s, a)\), so this approximation holds:
\[
\approx \frac{\partial f(s, a)}{\partial s}\bigg|_{(s_t, a_t)} \frac{d\mathcal{L}_\pi(\boldsymbol{\theta})}{d s_{t+1}}.
\]
Finally, this simplifies to:
\[
\frac{d s_{t+1}}{d s_t} \frac{d\mathcal{L}_\pi(\boldsymbol{\theta})}{d s_{t+1}} = \frac{d\mathcal{L}_\pi(\boldsymbol{\theta})}{d s_t}.
\]

The same principle applies to gradients with respect to actions:
\[
\frac{d \hat{s}_{t+1}}{d a_t} \frac{d\mathcal{L}_\pi(\boldsymbol{\theta})}{d \bar{s}_{t+1}} \approx \frac{\partial f(s, a)}{\partial a}\bigg|_{(s_t, a_t)} \frac{d\mathcal{L}_\pi(\boldsymbol{\theta})}{d s_{t+1}}.
\]
This simplifies further to:
\[
\frac{d s_{t+1}}{d a_t} \frac{d\mathcal{L}_\pi(\boldsymbol{\theta})}{d s_{t+1}} = \frac{d\mathcal{L}_\pi(\boldsymbol{\theta})}{d a_t}.
\]

This is the backpropagation formula of the analytical policy gradient. In summary, the \verb|GradientSwapingFunction| ensures that the simulator-derived \(s_{t+1}\) is used for accurate trajectory unrolling, while the gradients flow back through the learned dynamics model, evaluated at the accurate \(s_t\).

\subsubsection{Reparameterization Trick and Analytical Policy Gradient} \label{appdx:reparam_trick}
FoG-MBRL usually models $f(\cdot | s, a)$ and $\pi_\theta(\cdot | s)$ as Gaussian distributions with diagonal covariance: $f(\cdot | s, a)\sim\mathcal{N}\left(\mu(s, a), \Sigma(s, a)\right)$ and $\pi_\theta(\cdot | s)\sim\mathcal{N}\left(\mu_\theta(s), \Sigma_\theta(s)\right)$. This probabilistic representation serves dual purposes: introducing exploration through policy stochasticity and capturing aleatoric uncertainty \cite{chua2018deep} with model stochasticity. Resorting to Gaussian distributions enables the use of the reparameterization trick \cite{kingma2022autoencodingvariationalbayes}, where samples from these distributions are generated by sampling from a standard Gaussian $\mathcal{N}(0, 1)$, scaling by the covariance (e.g., $\Sigma(s, a)$ or $\Sigma_\theta(s)$), and shifting by the mean (e.g., $\mu(s, a)$ or $\mu_\theta(s)$). This trick preserves the original distribution while allowing gradients to flow through the sampling operation, thus allowing to directly optimize the RL objective given in \eqref{fo:mdp} with respect to $\theta$, with Stochastic Gradient Descent (SGD). This is in contrast to most MFRL methods that don't assume any specific form for the distributions of the dynamic function.

\subsection{Experimental Results}
\subsubsection{Detailed Results Across Environments}\label{appdx:detailed_results}

\begin{figure*}[h!]
    \centering
    \includegraphics[width=1.\textwidth]{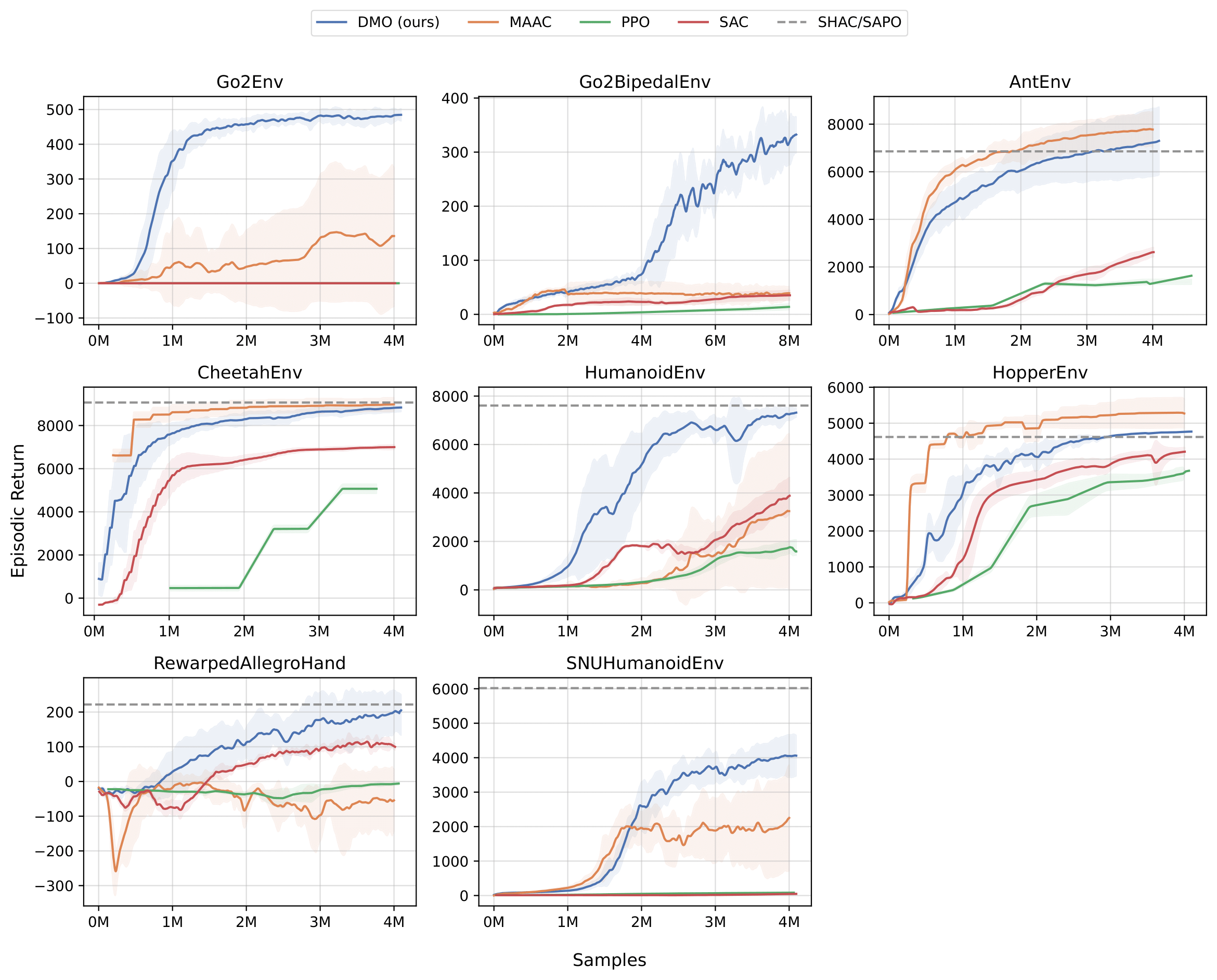}
    \caption{\textbf{Episodic Return Performance Across Environments.} This figure shows the episodic return (mean ± std) at 4M samples (8M for Go2BipedalEnv). DMO represents the best-performing DMO version for each environment. Although SHAC utilizes true model derivatives and does not rely on learned dynamics, DMO achieves competitive performance. SHAC results are unavailable for Go2Env and Go2BipedalEnv due to IsaacGym not being a differentiable simulator.}
    \label{fig:dflex_compare_indi_short}
\end{figure*}
In Figure \ref{fig:dflex_compare_indi_short} we also included the results of SHAC \cite{xu2021accelerated} (or SAPO \cite{xing2024stabilizing} for AllegroHand) for comparison, an algorithm close to ours but that uses the true derivatives instead of approximated ones. However, on the Go2 walking and Go2 bipedal motion environment, SHAC couldn't be trained because we lacked a differentiable simulator.

\subsection{Go2 Environment Specifications} 
These rewards use quantities not easily available on the real robot, such as the linear velocity of the robot's base and foot contact forces. The model has to learn these privileged measurements to provide a gradient for them, yet the policy cannot learn from these quantities if we want to deploy it on the real robot. Therefore, we used a different set of observations for the policy than for the dynamics model.
\subsubsection{Quadrupedal Task Reward and Observations}
\label{appdx:go2_quadrupedal_reward}

\begin{table*}[h!]
\centering
\caption{\textbf{List of Reward Terms and Formulas for the Go2 Quadrupedal Task.}}
\begin{tabular}{p{0.3\linewidth}p{0.65\linewidth}}
\toprule
\textbf{Reward Term} & \textbf{Formula} \\ 
\midrule

xy velocity tracking & $k_{\mathrm{xy\_vel}} \cdot e^{-\frac{\| \mathbf{v}_{xy} - \mathbf{v}_{xy}^\text{cmd} \|_2^2}{\sigma_{v_{xy}}}}$ \\ 
\addlinespace
yaw velocity tracking & $k_{\mathrm{yaw\_vel}} \cdot e^{-\frac{\left(\omega_z - \omega_z^\text{cmd}\right)^2}{\sigma_{\omega_z}}}$ \\ 
\addlinespace
z velocity penalty & $-k_{\mathrm{z\_vel}} \cdot \mathbf{v}_z^2$ \\ 
\addlinespace
roll-pitch velocity penalty & $-k_{\mathrm{rp\_vel}} \cdot \| \bm{\omega}_{xy} \|_2^2$ \\
\addlinespace
orientation penalty & $-k_{\mathrm{orient}} \cdot \| \text{ProjGravity}_{xy} \|_2^2$ \\
\addlinespace
action rate penalty & $-k_{\mathrm{act}} \cdot \| \mathbf{a}_{t-1} - \mathbf{a}_t \|^2_2$ \\
2nd order action rate penalty & $-k_{\mathrm{act}} \cdot \| (\mathbf{a}_{t-2} - \mathbf{a}_{t-1}) - (\mathbf{a}_{t-1} - \mathbf{a}_t) \|^2_2$ \\
\addlinespace
joint velocities penalty & $-k_{\mathrm{dotq}} \cdot \| \mathbf{\dot{q}} \|^2_2$ \\
\addlinespace
Raibert heuristic footswing tracking & $-k_{\mathrm{raibert}} \cdot \| \mathbf{p}_{x, y, \text{foot}}^{\text{ground}} - \mathbf{p}_{x, y, \text{raibert}}^{\text{ground}}(\mathbf{v}_{xy}^\text{cmd}, \omega_z^\text{cmd}) \|_2^2$ \\
\addlinespace
contact plan tracking & $-k_{\mathrm{cpt}} \cdot \sum_{\text{foot}} \left( 1 - C_{\text{foot}} \left( t \right) \right) \left(1 - e^{ -\frac{\lvert \mathbf{f}^{\text{foot}} \rvert^2}{\sigma_{c_f}}}\right)$ \\
\addlinespace
footswing height tracking & $-k_{\mathrm{fht}} \cdot \sum_{\text{foot}} \left( h_{z, \text{foot}}^{\text{ground}} - h_{z,\text{target},\text{foot}}^{\text{ground}} \right)^2 \left(1-C_{\text{foot}} \left( t \right)\right)$ \\

\bottomrule
\end{tabular}
\end{table*}

\medskip
\noindent\textbf{Definitions and Parameters:}
\begin{itemize}
  \item Scaling factors ($k_{\mathrm{...}}$ values):
  \begin{itemize}
    \item $k_{\mathrm{xy\_vel}} = 0.5$, $k_{\mathrm{yaw\_vel}} = 1.0$, $k_{\mathrm{z\_vel}} = 0.02$
    \item $k_{\mathrm{rp\_vel}} = 0.001$, $k_{\mathrm{orient}} = 5.0$, $k_{\mathrm{act}} = 0.1$
    \item $k_{\mathrm{dotq}} = 0.0001$, $k_{\mathrm{raibert}} = 10.0$, $k_{\mathrm{cpt}} = 1.0$
    \item $k_{\mathrm{fht}} = 30.0$
  \end{itemize}
  \item Key variables:
  \begin{itemize}
    \item $h^{\mathrm{ground}}_{z,\text{foot}}$: Height of foot $f$ from ground plane
    \item $h^{\mathrm{ground}}_{z,\mathrm{target},\text{foot}}(t)$: Target foot height from gait generator
    \item $C_{\text{foot}}(t)$: Desired contact state (0=swing, 1=stance)
  \end{itemize}
\end{itemize}

\begin{table}[h!] 
\centering
\caption{\textbf{List of Observations for the Dynamics Model and Actor in the Go2 Quadrupedal Task.}}
\begin{tabular}{p{0.38\linewidth}p{0.17\linewidth}p{0.17\linewidth}p{0.17\linewidth}}
\toprule
\textbf{Observation} & \textbf{Size} & \textbf{Dyn. model} & \textbf{Actor} \\ 
\midrule

linear velocity of the base & $3$ & \checkmark & \\ 
\addlinespace
angular velocity of the base & $3$ & \checkmark & \checkmark \\ 
\addlinespace
command & $3$ & \checkmark & \checkmark \\ 
\addlinespace
projected gravity & $3$ & \checkmark & \checkmark \\ 
\addlinespace
joint positions & $12$ & \checkmark &  \checkmark\\ 
\addlinespace
joint velocities & $12$ & \checkmark & \checkmark \\ 
\addlinespace
previous actions & $12$ & \checkmark & \checkmark \\ 
\addlinespace
clock inputs & $4$ & \checkmark & \checkmark \\ 
\addlinespace
foot x and y positions & $8$ & \checkmark & \\ 
\addlinespace
foot forces & $4$ & \checkmark & \\ 
\addlinespace
foot heights & $4$ & \checkmark & \\ 
\bottomrule

\end{tabular}
\end{table}

\clearpage
\subsubsection{Bipedal Task Reward and Observations}
\label{appdx:go2_bipedal_reward}
\begin{table*}[h!]
\centering
\caption{\textbf{Reward Terms and Formulas for the Go2 Bipedal Task.}}
\begin{tabular}{p{0.30\linewidth}p{0.65\linewidth}}
\toprule
\textbf{Reward Term} & \textbf{Formula} \\
\midrule

No Velocity Reward/Penalty & 
$\displaystyle k_{\mathrm{lin}} \exp\!\Bigl(-\tfrac{\|\mathbf{v}_{xy}\|^{2}}{\delta_{\mathrm{lin}}}\Bigr)
\cdot \mathbf{1}_{\mathrm{stand}}
\cdot \tfrac{\mathrm{clip}(h,h_{\min},h_{\max}) - h_{\min}}{h_{\max}-h_{\min}}
\;-\;k_{\mathrm{vel\_pen}} \|\mathbf{v}_{xy}\|^{2}\mathbf{1}_{t > t_{v}}$ \\[1.5ex]

Angular Velocity Penalty & 
$-k_{\omega} \|\bm{\omega}_{z}\|^{2} \mathbf{1}_{t>t_{c}}$ \\[1ex]

Action Rate Penalty &
$-k_{\mathrm{act}} \| \mathbf{a}_t - \mathbf{a}_{t-1} \|_2^2$ \\[1ex]

2nd Order Action Rate Penalty &
$-k_{\mathrm{act}} \| (\mathbf{a}_{t-2} - \mathbf{a}_{t-1}) - (\mathbf{a}_{t-1} - \mathbf{a}_t) \|_2^2$ \\[1ex]

Joint Velocities Penalty & 
$-k_{\dot{q}} \| \mathbf{\dot{q}} \|_2^2$ \\[1ex]

Torque Limits Penalty & 
$-k_{\tau} \sum_{j}\max\bigl(0,|\tau_{j}|-\tau_{\max}\sigma_{s}\bigr)$ \\[1ex]

Footswing Height Tracking & 
$-k_{\mathrm{clr}} \mathbf{1}_{t > t_c} \sum\limits_{f \in F_{\mathrm{front}}} \bigl(h^{\mathrm{ground}}_{z,f} - h^{\mathrm{ground}}_{z,\mathrm{target},f}\bigr)^2 (1 - C_{f}(t))$ \\[1ex]

Contact Plan Tracking & 
$-k_{\mathrm{cfs}} \mathbf{1}_{\mathrm{stand}} \sum\limits_{f \in F_{\mathrm{front}}} (1 - C_{f}(t)) \bigl(1 - e^{-\|\mathbf{f}_{f}\|^{2}/\sigma_{c_f}}\bigr)$ \\[1ex]

Stand-Air Penalty/Reward & 
$\mathbf{1}_{t < t_c}\Bigl(
  -k_{\mathrm{air\_pen}} \sum\limits_{f \in F_{\mathrm{front}}}\max(0,h^{\mathrm{ground}}_{z,f}-0.06)
  \;+\;k_{\mathrm{air\_rew}} \sum\limits_{f \in F_{\mathrm{back}}}\min(h^{\mathrm{ground}}_{z,f},0.06)
\Bigr)$ \\[1ex]

Lift-Up Reward & 
$k_{\mathrm{lift}} \cdot \mathrm{clip}\!\Bigl(\tfrac{H - H_{\min}}{H_{\max}-H_{\min}},\,0,\,1\Bigr)$ \\[1ex]

Upright Posture Reward & 
$k_{\mathrm{up}} \cdot \Bigl(0.5 \cdot \tfrac{\mathbf{v}_f \cdot \mathbf{v}_u}{\|\mathbf{v}_u\|} + 0.5\Bigr)^2$ \\

\bottomrule
\end{tabular}
\end{table*}

\medskip
\noindent\textbf{Definitions and Parameters:}
\begin{itemize}
  \item $\mathbf{1}_{\mathrm{stand}} = \begin{cases} 
  1 & \text{if } \tfrac{\mathbf{v}_f \cdot \mathbf{v}_u}{\|\mathbf{v}_u\|} > 0.9 \\
  0 & \text{otherwise}
  \end{cases}$
  
  \item Scaling factors ($k_{\mathrm{...}}$ values):
  \begin{itemize}
    \item $k_{\mathrm{lin}} = 1.0$, $k_{\mathrm{vel\_pen}} = 0.4$, $k_{\omega} = 0.1$
    \item $k_{\mathrm{act}} = 0.03$, $k_{\dot{q}} = 0.0001$, $k_{\tau} = 0.01$
    \item $k_{\mathrm{clr}} = 300.0$, $k_{\mathrm{cfs}} = 1.0$, $k_{\mathrm{air\_pen}} = 40$, $k_{\mathrm{air\_rew}} = 5$
    \item $k_{\mathrm{lift}} = 0.5$, $k_{\mathrm{up}} = 1.0$
  \end{itemize}
  
  \item Vectors:
  \begin{itemize}
    \item $\mathbf{v}_u = R_z(\theta)\begin{bmatrix}0.2\\0\\-1.0\end{bmatrix}$ (yaw-rotated world vector)
    \item $\mathbf{v}_f = R_{WB}\begin{bmatrix}1.0\\0\\0\end{bmatrix}$ (robot's forward axis in world frame)
  \end{itemize}
  
  \item Key variables:
  \begin{itemize}
    \item $h^{\mathrm{ground}}_{z,f}$: Height of foot $f$ from ground plane
    \item $h^{\mathrm{ground}}_{z,\mathrm{target},f}(t)$: Target foot height from gait generator
    \item $C_{f}(t)$: Desired contact state (0=swing, 1=stance)
    \item $H$: Robot base height, $\tau_{\max}$: Torque limits
    \item $\sigma_{s} = 0.5$: Torque limit safety margin
  \end{itemize}
\end{itemize}

\begin{table*}[h!] 
\centering
\caption{\textbf{List of Observations for the Dynamics Model and Actor in the Go2 Bipedal Task.}}
\begin{tabular}{p{0.38\linewidth}p{0.17\linewidth}p{0.17\linewidth}p{0.17\linewidth}}
\toprule
\textbf{Observation} & \textbf{Size} & \textbf{Dyn. model} & \textbf{Actor} \\ 
\midrule

linear velocity of the base & $3$ & \checkmark & \\ 
\addlinespace
angular velocity of the base & $3$ & \checkmark & \checkmark \\ 
\addlinespace
projected gravity & $3$ & \checkmark & \checkmark \\ 
\addlinespace
joint positions & $12$ & \checkmark &  \checkmark\\ 
\addlinespace
joint velocities & $12$ & \checkmark & \checkmark \\ 
\addlinespace
previous actions & $12$ & \checkmark & \checkmark \\ 
\addlinespace
clock inputs & $4$ & \checkmark & \checkmark \\ 
\addlinespace
joint torques & $12$ & \checkmark & \\ 
\addlinespace
foot forces & $4$ & \checkmark & \\ 
\addlinespace
foot heights & $4$ & \checkmark & \\ 
\addlinespace
$\mathbf{v}_f$ & $3$ & \checkmark & \\ 
\addlinespace
$\mathbf{v}_u$ & $3$ & \checkmark & \\ 
\addlinespace
base height & $1$ & \checkmark & \\ 
\bottomrule

\end{tabular}
\end{table*}

\clearpage
\subsection{Hyperparameters}
\begin{table*}[h!]
\centering
\caption{\textbf{Shared Hyperparameters.} Algorithms use these settings unless otherwise specified in the environment-specific tables below.}
\begin{tabular}{lccccc}
\toprule
& PPO & SAC & DMO-BPTT & DMO-SHAC & DMO-SAPO \\
\midrule
Num actors & * & * & 256 & * & * \\
Horizon & 32 & 32 & 16 & 16 & 16 \\
Mini-epochs & 5 & 8 & 1 & 16 & 16 \\
Discount $\gamma$ & 0.99 & 0.99 & 0.99 & 0.99 & 0.99 \\
TD/GAE $\lambda$ & 0.95 & 0.95 & -- & 0.95 & 0.95 \\
Actor lr & $3\mathrm{e}{-4}$ & * & $3\mathrm{e}{-4}$ & * & $2\mathrm{e}{-3}$ \\
Critic lr & $3\mathrm{e}{-4}$ & * & $5\mathrm{e}{-4}$ & * & $5\mathrm{e}{-4}$ \\
Dyn. model lr & -- & -- & * & * & $3\mathrm{e}{-4}$ \\
Entropy lr & -- & * & -- & -- & $5\mathrm{e}{-3}$ \\
lr schedule & KL(0.008) & -- & linear & linear & linear \\
Optim type & AdamW & Adam & Adam & Adam & AdamW \\
Optim $(\beta_1, \beta_2)$ & (0.9, 0.999) & (0.7, 0.95) & (0.7, 0.95) & (0.7, 0.95) & (0.7, 0.95) \\
Grad clip & 0.5 & 0.5 & 1.0 & 1.0 & 1.0 \\
Norm type & LayerNorm & LayerNorm & LayerNorm & LayerNorm & LayerNorm \\
Act type & ELU & ELU & ELU & ELU & SiLU \\
Actor $\sigma(s)$ & yes & no & no & no & yes \\
Num critics & -- & 2 & -- & -- & * \\
Critic $\tau$ & -- & 0.995 & -- & * & -- \\
Replay buffer & -- & $10^6$ & -- & $10^6$ & $10^6$ \\
Target entropy $\mathcal{H}$ & -- & $-\dim(\mathcal{A})/2$ & -- & -- & $-\dim(\mathcal{A})/2$ \\
Init temperature & -- & 1.0 & -- & -- & 1.0 \\
\bottomrule
\multicolumn{6}{l}{\footnotesize * Values vary by environment, see environment-specific tables.}
\end{tabular}
\end{table*}

\begin{table}[h!]
\centering
\caption{\textbf{Environment-specific Hyperparameters for PPO.}}
\resizebox{\columnwidth}{!}{%
\begin{tabular}{lcccc}
\toprule
\textbf{Environment} & \textbf{Num actors} & \textbf{Minibatch size} & \textbf{Actor MLP} & \textbf{Critic MLP} \\
\midrule
Ant & 2048 & 16384 & (128,64,32) & (128,64,32) \\
Hopper/Cheetah & 1024 & 8192 & (128,64,32) & (128,64,32) \\
Humanoid & 1024 & 8192 & (256,128,64) & (256,128,64) \\
SNUHumanoid & 1024 & 8192 & (512,512,256) & (512,512,256) \\
Go2/Go2Bipedal & 4096 & 32768 & (256,128,64) & (256,128,64) \\
AllegroHand & 4096 & 32768 & (400,400,200,100) & (400,400,200,100) \\
\bottomrule
\end{tabular}%
}
\end{table}

\begin{table}[h!]
\centering
\caption{\textbf{Environment-specific Hyperparameters for SAC.}}
\resizebox{\columnwidth}{!}{%
\begin{tabular}{lccccccc}
\toprule
\textbf{Environment} & \textbf{Actors} & \textbf{Batch} & \textbf{Actor lr} & \textbf{Critic lr} & \textbf{Entropy lr} & \textbf{Actor MLP} & \textbf{Critic MLP} \\
\midrule
Ant & 128 & 4096 & $5\mathrm{e}{-4}$ & $5\mathrm{e}{-4}$ & $5\mathrm{e}{-3}$ & (256,128,64) & (256,128,64) \\
Hopper/Cheetah & 64 & 2048 & $5\mathrm{e}{-4}$ & $5\mathrm{e}{-4}$ & $5\mathrm{e}{-3}$ & (256,128,64) & (256,128,64) \\
Humanoid & 64 & 2048 & $3\mathrm{e}{-4}$ & $3\mathrm{e}{-4}$ & $2\mathrm{e}{-4}$ & (512,256) & (512,256) \\
SNUHumanoid & 256 & 4096 & $3\mathrm{e}{-4}$ & $3\mathrm{e}{-4}$ & $2\mathrm{e}{-4}$ & (512,512,512,256) & (512,512,512,256) \\
Go2/Go2Bipedal/AllegroHand & 64 & 2048 & $3\mathrm{e}{-4}$ & $3\mathrm{e}{-4}$ & $2\mathrm{e}{-4}$ & (512,256) & (512,256) \\
\bottomrule
\end{tabular}%
}
\end{table}

\begin{table}[h!]
\centering
\caption{\textbf{Environment-specific Hyperparameters for DMO-SHAC.}}
\resizebox{\columnwidth}{!}{%
\begin{tabular}{lccccccc}
\toprule
\textbf{Environment} & \textbf{Actors} & \textbf{Actor lr} & \textbf{Critic lr} & \textbf{Dyn. lr} & \textbf{Critic $\tau$} & \textbf{Actor MLP} & \textbf{Dyn. Model MLP} \\
\midrule
Ant/Cheetah & 64 & $2\mathrm{e}{-3}$ & $2\mathrm{e}{-3}$ & $2\mathrm{e}{-3}$ & 0.2 & (128,64,32) & (512,512) \\
Hopper & 256 & $2\mathrm{e}{-3}$ & $2\mathrm{e}{-4}$ & $2\mathrm{e}{-3}$ & 0.2 & (128,64,32) & (512,512) \\
Humanoid & 64 & $2\mathrm{e}{-3}$ & $5\mathrm{e}{-4}$ & $3\mathrm{e}{-4}$ & 0.995 & (256,128) & (1792,1792) \\
SNUHumanoid & 64 & $2\mathrm{e}{-3}$ & $5\mathrm{e}{-4}$ & $3\mathrm{e}{-4}$ & 0.995 & (512,256) & (1792,1792) \\
Go2 & 256 & $3\mathrm{e}{-4}$ & $5\mathrm{e}{-4}$ & $3\mathrm{e}{-4}$ & 0.2 & (256,128,64) & (1024,1024) \\
\bottomrule
\end{tabular}%
}
\end{table}

\begin{table}[h!]
\centering
\caption{\textbf{DMO-SAPO Environment-specific Hyperparameters.}}
\begin{tabular}{lccccc}
\toprule
\textbf{Environment} & \textbf{Actors} & \textbf{Num Critics} & \textbf{Actor MLP} & \textbf{Dyn. Model MLP} \\
\midrule
Go2Bipedal & 512 & 10 & (256,128,64) & (1792,1792) \\
AllegroHand & 128 & 2 & (512,256) & (1792,1792) \\
\bottomrule
\end{tabular}
\end{table}

\paragraph{Hyperparameter Sources.}
Hyperparameters for PPO, SAC, and the SHAC part of DMO-SHAC for Ant, Cheetah, Hopper, Humanoid, and SNUHumanoid were taken from \citep{xu2021accelerated}. Hyperparameters for PPO and SAC for AllegroHand were taken from \citep{makoviychuk2021isaac}. Hyperparameters for the SAPO part of DMO-SAPO for the AllegroHand environment were from \citep{xing2024stabilizing}.

\subsection{Comparative Summary of First-order Gradient Algorithms}
\begin{table}[h]
\centering
\caption{Comparison of first-order gradient RL algorithms along key features.}
\begin{tabular}{lcccccc}
\toprule
Algorithm & Needs no diff. sim. & Value bootstrap & Parallel sim. & Decoupling \\
\midrule
SVG($\infty$)      & \cmark & \xmark & \xmark & \cmark \\
BPTT               & \xmark & \xmark & \cmark & \xmark \\
SHAC/SAPO          & \xmark & \cmark & \cmark & \xmark \\
PWM                & \cmark & \cmark & \cmark & \xmark \\
MAAC/Dreamer       & \cmark & \cmark & \xmark & \xmark \\
DMO                & \cmark & \cmark & \cmark & \cmark \\
\bottomrule
\end{tabular}
\end{table}

\end{document}